\documentclass[draftclsnofoot,onecolumn]{IEEEtran}
\usepackage{cite,graphicx,float,url,amsmath,amssymb,amsthm,multirow,algorithmic}
\usepackage[tight,footnotesize]{subfigure}
\usepackage[boxed,noresetcount,ruled,vlined]{algorithm2e}
\usepackage[usenames]{color}

\graphicspath{{Figures/}}

\newcommand{\cB}{\mathcal{B}}

\newcommand{\cC}{\mathcal{C}}

\newcommand{\cL}{\mathcal{L}}

\newcommand{\R}{\mathbb{R}}

\newcommand{\cS}{\mathcal{S}}

\newcommand{\cT}{\mathcal{T}}

\newcommand{\cU}{\mathcal{U}}

\DeclareMathOperator*{\argmin}{arg\,min}

\newcommand{\diag}{\mathrm{diag}}

\newcommand{\tr}{\mathrm{tr}}
\newcommand{\rank}{\mathrm{rank}}

\begin{document}
%
\title{Human Action Attribute Learning From Video Data Using Low-Rank Representations}
%
%
%

\author{Tong~Wu,~\IEEEmembership{Student Member,~IEEE,}
        Prudhvi~Gurram,~\IEEEmembership{Senior~Member,~IEEE,} \\
        Raghuveer~M.~Rao,~\IEEEmembership{Fellow,~IEEE,}
        and~Waheed~U.~Bajwa,~\IEEEmembership{Senior~Member,~IEEE}
\thanks{This work is supported in part by the NSF under grants CCF-1218942 and CCF-1453073, by the Army Research Office under grant W911NF-14-1-0295, and by an Army Research Lab Robotics CTA subaward. Preliminary versions of parts of this work have been presented at the 2016 IEEE Image Video and Multidimensional Signal Processing (IVMSP) workshop \cite{WuGRB.IVMSP2016}.}
\thanks{Tong Wu and Waheed U. Bajwa are with the Department of Electrical and Computer Engineering, Rutgers University, Piscataway, NJ 08854, USA (E-mails: \{tong.wu.ee, waheed.bajwa\}@rutgers.edu). Prudhvi Gurram is with Booz Allen Hamilton, McLean, VA 22102, and also with U.S. Army Research Laboratory, Adelphi, MD 20783, USA (Email: pkgurram@ieee.org). Raghuveer M. Rao is with U.S. Army Research Laboratory, Adelphi, MD 20783, USA (Email: raghuveer.m.rao.civ@mail.mil).}}

\maketitle

\begin{abstract}
Representation of human actions as a sequence of human body movements or action attributes enables the development of models for human activity recognition and summarization. We present an extension of the low-rank representation (LRR) model, termed the clustering-aware structure-constrained low-rank representation (CS-LRR) model, for unsupervised learning of human action attributes from video data. Our model is based on the union-of-subspaces (UoS) framework, and integrates spectral clustering into the LRR optimization problem for better subspace clustering results. We lay out an efficient linear alternating direction method to solve the CS-LRR optimization problem. We also introduce a hierarchical subspace clustering approach, termed hierarchical CS-LRR, to learn the attributes without the need for a priori specification of their number. By visualizing and labeling these action attributes, the hierarchical model can be used to semantically summarize long video sequences of human actions at multiple resolutions. A human action or activity can also be uniquely represented as a sequence of transitions from one action attribute to another, which can then be used for human action recognition. We demonstrate the effectiveness of the proposed model for semantic summarization and action recognition through comprehensive experiments on five real-world human action datasets.
\end{abstract}


%
\IEEEpeerreviewmaketitle

\section{Introduction}

\IEEEPARstart{T}{he} need for high-level analytics of large streams of video data has arisen in recent years in many practical commercial, law enforcement, and military applications \cite{NiuLHW.ICME2004,AggarwalRK.Totorial2011,JiangBCS.IJMIR2013,RyooM.CVPR2013}. Examples include human activity recognition, video summarization and indexing, human-machine teaming, and human behavior tracking and monitoring. The human activity recognition process requires recognizing both objects in the scene as well as body movements to correctly identify the activity with the help of context \cite{ZhuNR.CVPR2013}. In the case of video summarization, one should be able to semantically segment and summarize the visual content in terms of context. This enables efficient indexing of large amounts of video data, which allows for easy query and retrieval \cite{HuXLZM.SMC2011}. For effective human-machine (robot) teaming, autonomous systems should be able to understand human teammates' gestures as well as recognize various human activities taking place in the field to gain situational awareness of the scene. Further, an important step in building visual tracking systems is to design ontologies and vocabularies for human activity and environment representations \cite{AkdemirTC.MM2008,RodriguezCLC.CSUR2014}. All these tasks necessitate automated learning of movements of the human body or human action attributes.

\begin{figure}[t]
\centering
\includegraphics[width=3in]{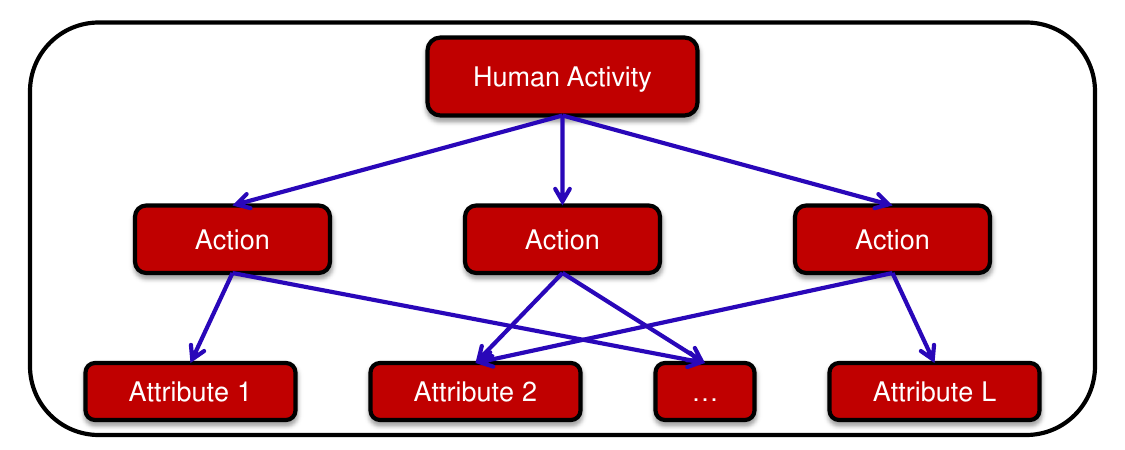}
\caption{The hierarchical model for complex human activity representation.}
\label{fig:hiemodel}
\end{figure}

Human activities consist of a sequence of actions that can be represented hierarchically \cite{JiangBCS.IJMIR2013} as shown in Fig.~\ref{fig:hiemodel}. The bottom level of the hierarchy consists of the fine resolution description of an action, i.e., movement of the human body (e.g., right arm moves up, left arm moves up, torso bending, and legs moving apart) and can be called an action attribute \cite{LiuKS.CVPR2011}. At the middle level, a sequence of these attributes forms a human action. The human actions and their interactions form an activity, while a sequence of activities forms an event. An important advantage of the hierarchical model is that such structures go hand in hand with semantic or syntactic approaches and they provide us with the flexibility to generate summaries of long video sequences at different resolutions based on the needs of the end application.

In this work, we focus on the bottom two layers of the human activity hierarchical model, i.e., human actions and their representations using attributes. One possible way of obtaining such representations is to manually specify the action attributes and assign training video sequences to each attribute \cite{LiuKS.CVPR2011}. Another way is to manually annotate training video sequences by labeling movements \cite{RamananF.NIPS2003}. Any human action in a test sequence can then be described using such user-defined attributes. Both of these approaches fall into supervised category. However, a set of user-defined action attributes may not completely describe all the human actions in given data. Also, manual assignment of training data for each of the action attributes is time consuming, if not impossible, for large datasets. Another issue with supervised learning of action attributes is that there might be attributes that were not seen in the training data by the system, but that might be seen in the field. Because of these reasons, unsupervised techniques to learn action attributes have been investigated \cite{LiuS.CVPR2008,LiuYS.CVPR2009,YuanYW.CVPR2011}. These methods learn action attributes by clustering low-level features based on their co-occurrence in training videos. However, video data are not usually well distributed around cluster centers and hence, the cluster statistics may not be sufficient to accurately represent the attributes \cite{ElhamifarV.PAMI2013}.

Motivated by the premise that high-dimensional video data usually lie in a union of low-dimensional subspaces, instead of being uniformly distributed in the high-dimensional ambient space \cite{ElhamifarV.PAMI2013}, we propose to represent human action attributes based on the \emph{union-of-subspaces} (UoS) model \cite{LiuLYSYM.PAMI2013}. The hypothesis of the UoS model is that each action attribute can be represented by a subspace. We conjecture that the action attributes represented by subspaces can encode more variations within an attribute compared to the representations obtained using co-occurrence statistics \cite{LiuS.CVPR2008,LiuYS.CVPR2009,YuanYW.CVPR2011}. The task of unsupervised learning of the UoS underlying data of interest is often termed \emph{subspace clustering} \cite{LuMZZHY.ECCV2012,ElhamifarV.PAMI2013,LiuLYSYM.PAMI2013}, which involves learning a graph associated with the data and then applying spectral clustering on the graph to infer the clustering of data. Recently, the authors in \cite{ElhamifarV.PAMI2013} have developed a sparse subspace clustering (SSC) technique by solving an $\ell_1$-minimization problem. This has been extended into a hierarchical structure to learn subspaces at multiple resolutions in \cite{WuGRB.ICCVW2015}. To capture the global structure of data, low-rank representation (LRR) models with and without sparsity constraints have been proposed in \cite{ZhuangGTWLMY.TIP2015} and \cite{LiuLYSYM.PAMI2013}, respectively. Liu \emph{et al.} \cite{LiuCZX.TIP2014} extended LRR by incorporating manifold regularization into the LRR framework. It has been proved that LRR can achieve perfect subspace clustering results under the condition that the subspaces underlying the data are independent \cite{LiuLY.ICML2010,LiuLYSYM.PAMI2013}. However, this condition is hard to satisfy in many real situations. To handle the case of disjoint subspaces, Tang \emph{et al.} \cite{TangLSZ.TNNLS2014} extended LRR by imposing restrictions on the structure of the solution, called structure-constrained LRR (SC-LRR). The low-rank subspace learning has been extended to multidimensional data for action recognition \cite{JiaF.TIP2016}. However, there are fundamental differences between our work and \cite{JiaF.TIP2016} because the main objective of our proposed approach is to learn action attributes in an unsupervised manner.

\subsection{Our Contributions}
\label{ssec:contribute}

Existing LRR based subspace clustering techniques use spectral clustering as a post-processing step on the graph generated from a low-rank coefficient matrix, but the relationship between the coefficient matrix and the segmentation of data is seldom considered, which can lead to sub-optimal results \cite{GaoNLH.ICCV2015}. Our first main contribution in this regard is introduction of a novel low-rank representation model, termed \emph{clustering-aware structure-constrained LRR} (CS-LRR) model, to obtain optimal clustering of human action attributes from a large collection of video sequences. We formulate the CS-LRR learning problem by introducing spectral clustering into the optimization program. The second main contribution of this paper is a hierarchical extension of our CS-LRR model for unsupervised learning of human action attributes from the data at different resolutions without assuming any knowledge of the number of attributes present in the data. Once the graph is learned from CS-LRR model, we segment it by applying hierarchical spectral clustering to obtain action attributes at different resolutions. The proposed approach is called \emph{hierarchical clustering-aware structure-constrained LRR} (HCS-LRR).

\begin{figure}[t]
\centering
\includegraphics[height=1.8in]{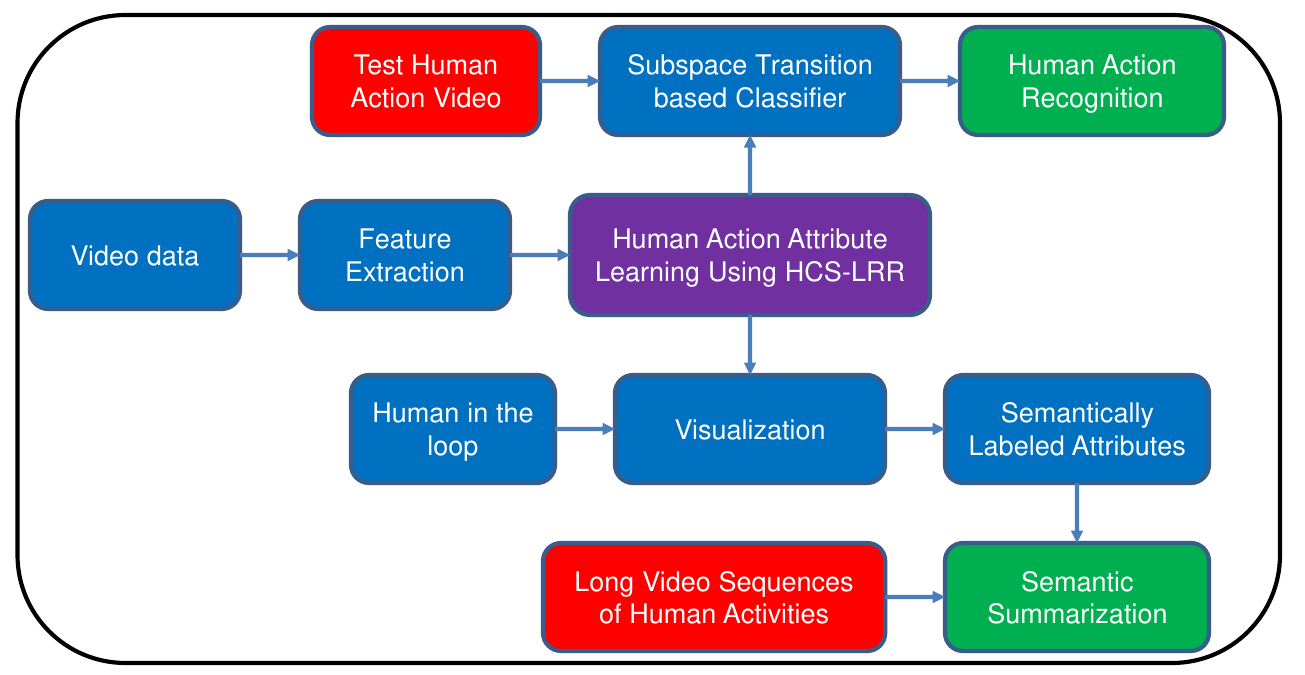}
\caption{Block diagram of the human activity recognition and summarization system using HCS-LRR algorithm.}
\label{fig:bldiag}
\end{figure}

The block diagram of the system that uses HCS-LRR algorithm to learn human action attributes is shown in Fig.~\ref{fig:bldiag}. A large stream of video data is taken as input and features such as silhouettes, frame-by-frame spatial features like histograms of oriented gradients (HOG) \cite{DalalT.CVPR2005}, and spatio-temporal features like motion boundary histogram (MBH) \cite{DalalTS.ECCV2006} are extracted from the input. The data samples in this high-dimensional feature space are given as the input to the HCS-LRR algorithm and human action attributes are obtained as the output. One of the main applications of learning the attributes based on the UoS model is \emph{semantic summarization} of long video sequences. The attributes at different levels of the hierarchy can be labeled by an expert-in-the-loop by visualizing the first few basis vectors of each attribute (subspace) in the form of images. Once the labeled attributes are available, any long video sequence of human activity can then be semantically summarized at different levels of granularity based on the requirements of an application. Another major application of learning the attributes is \emph{human action recognition}. A human action or activity can be represented as a sequence of transitions from one attribute to another, and hence, can be represented by a subspace transition vector. Even though multiple actions can share action attributes, each action or activity can be uniquely represented by its subspace transition vector. A classifier can be trained based upon these transition vectors to classify an action in a test video sequence into one of the actions in the training data. Our final contribution involves developing frameworks for both semantic summarization and human action recognition from the HCS-LRR model. Our results confirm the superiority of HCS-LRR in comparison to a number of state-of-the-art subspace clustering approaches.

\subsection{Notational Convention and Organization}
\label{ssec:notaandorg}

The following notation will be used throughout the rest of this paper. We use non-bold letters to represent scalars, bold lowercase letters to denote vectors/sets, and bold uppercase letters to denote matrices. The $i$-th element of a vector $\mathbf{a}$ is denoted by $a_{(i)}$ and the $(i,j)$-th element of a matrix $\mathbf{A}$ is denoted by $a_{i,j}$. The $i$-th row and $j$-th column of a matrix $\mathbf{A}$ are denoted by $\mathbf{a}^i$ and $\mathbf{a}_j$, respectively. Given two sets $\mathbf{\Omega}_1$ and $\mathbf{\Omega}_2$, $[\mathbf{A}]_{\mathbf{\Omega}_1, \mathbf{\Omega}_2}$ denotes the submatrix of $\mathbf{A}$ corresponding to the rows and columns indexed by $\mathbf{\Omega}_1$ and $\mathbf{\Omega}_2$, respectively. The zero matrix and the identity matrix are denoted by $\mathbf{0}$ and $\mathbf{I}$ of appropriate dimensions, respectively.

The only used vector norm in this paper is the $\ell_2$ norm, which is represented by $\|\cdot\|_2$. We use a variety of norms on matrices. The $\ell_1$ and $\ell_{2,1}$ norms are denoted by $\|\mathbf{A}\|_1 = \sum_{i,j} |a_{i,j}|$ and $\|\mathbf{A}\|_{2,1} = \sum_j \|\mathbf{a}_j\|_2$, respectively. The $\ell_{\infty}$ norm is defined as $\|\mathbf{A}\|_{\infty} = \max_{i,j} |a_{i,j}|$. The spectral norm of a matrix $\mathbf{A}$, i.e., the largest singular value of $\mathbf{A}$, is denoted by $\|\mathbf{A}\|$. The Frobenius norm and the nuclear norm (the sum of singular values) of a matrix $\mathbf{A}$ are denoted by $\|\mathbf{A}\|_F$ and $\|\mathbf{A}\|_{\ast}$, respectively. Finally, the Euclidean inner product between two matrices is $\langle \mathbf{A}, \mathbf{B} \rangle = \tr(\mathbf{A}^T \mathbf{B})$, where $(\cdot)^T$ and $\tr(\cdot)$ denote transpose and trace operations, respectively.

The rest of the paper is organized as follows. Section~\ref{sec:featureextract} introduces our feature extraction approach for human action attribute learning. In Section~\ref{sec:cslrralgo}, we mathematically formulate the CS-LRR model and present the algorithm based on CS-LRR model. The hierarchical structure of the CS-LRR model is described in Section~\ref{sec:hcslrr}. In Section~\ref{sec:semantic} and \ref{sec:recogmethod}, we discuss the approaches for semantic description of long video sequences and action recognition using learned action attributes, respectively. We then present experimental results in Section~\ref{sec:experiment}, which is followed by concluding remarks in Section~\ref{sec:conclude}.

\section{Feature Extraction for Attribute Learning}
\label{sec:featureextract}

\begin{figure*}[t]
\centering
\subfigure[]{\includegraphics[width=2.8in]{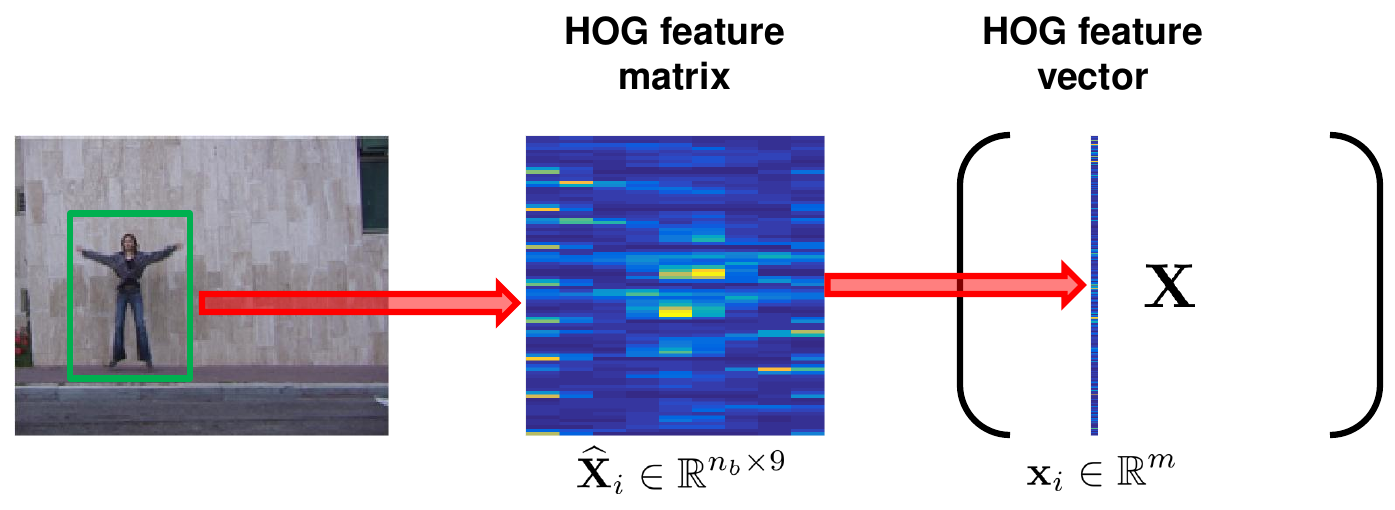} \label{fig:HOGexample}}
\qquad
\subfigure[]{\includegraphics[width=3.5in]{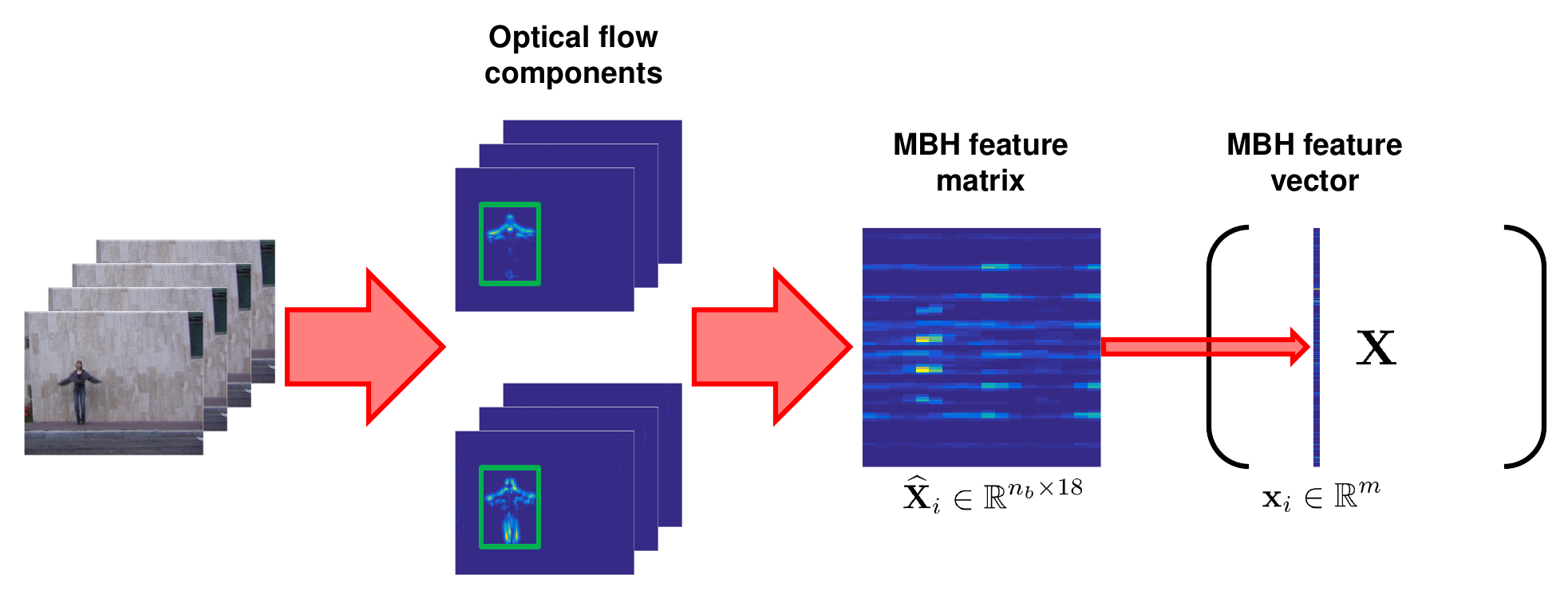} \label{fig:MBHexample}}
\caption{Illustration of our approach to create the matrix $\mathbf{X}$ for (a) HOG and (b) MBH features, which is then input to the HCS-LRR algorithm.}
\label{fig:featextract}
\end{figure*}

The main focus of our work is to learn meaningful human action attributes from large streams of video data in an unsupervised manner. The first step in our proposed framework is to extract feature descriptors from an action interest region in which the human performs the action. The action interest region of each frame of an action sequence is determined by a bounding box. In our work, we learn action attributes using two local visual descriptors: HOG (histograms of oriented gradients) \cite{DalalT.CVPR2005} and MBH (motion boundary histogram) \cite{DalalTS.ECCV2006}. To extract HOG descriptors, we divide the action interest region into a grid of blocks, each of size $n_{\sigma} \times n_{\sigma}$ pixels. Then HOG feature is extracted for each block and orientations are quantized into 9 bins. Therefore, the HOG feature of every frame can be stored into a matrix $\widehat{\mathbf{X}}_i \in \R^{n_b \times 9}$, where $n_b$ denotes the number of blocks in the action interest region, and the HOG feature vector of each block corresponds to a row in $\widehat{\mathbf{X}}_i$. We vectorize HOG features and normalize each vector to unit $\ell_2$ norm, forming individual data samples in a matrix $\mathbf{X} \in \R^{m \times N}$, where $m = n_b \times 9$ and $N$ denotes the total number of frames in the videos, as shown in Fig.~\ref{fig:HOGexample}.

The MBH descriptor represents the oriented gradients computed separately from the vertical and horizontal components of the optical flow, which is robust to camera and background motion. To extract MBH descriptors, we first split the optical flow field into two scalar fields corresponding to the horizontal and vertical components, which can be regarded as ``images'' of the motion components. Similar to HOG descriptor extraction, we divide the action interest region of each of the optical flow component image into a grid of blocks, each of size $n_{\sigma} \times n_{\sigma}$ pixels. Spatial derivatives are then computed for each block in each optical flow component and orientation information is quantized into 9 bins. Instead of using MBH features of optical flow field between every two video frames separately, we aggregate MBH features of every $n_{\tau}$ adjacent optical flow fields (computed between $n_{\tau} + 1$ video frames) and the sum is used as the feature of these $n_{\tau}$ optical flow fields. Therefore, the MBH feature of every $n_{\tau}$ adjacent optical flow fields corresponds to a matrix $\widehat{\mathbf{X}}_i \in \R^{n_b \times 18}$, where $n_b$ denotes the number of blocks in the action interest region, and the MBH feature vector of each block corresponds to a row in $\widehat{\mathbf{X}}_i$. We again vectorize the MBH features and normalize each vector to unit $\ell_2$ norm, forming MBH feature matrix $\mathbf{X} \in \R^{m \times N}$, where $m = n_b \times 18$ and $N$ again denotes the total number of feature descriptors, see Fig.~\ref{fig:MBHexample}. Given all the features extracted from the video data, we aim to learn action attributes based on the UoS model, which will be described in the following sections.

\section{Clustering-Aware Structure-Constrained Low-Rank Representation}
\label{sec:cslrralgo}

In this section, we propose our clustering-aware structure-constrained LRR (CS-LRR) model for learning the action attributes using the feature descriptors extracted from the video data. We begin with a brief review of LRR and SC-LRR since our CS-LRR model extends these models.

\subsection{Brief Review of LRR and SC-LRR}
\label{ssec:review}

Consider a collection of $N$ feature vectors in $\R^m$, $\mathbf{X} = [\mathbf{x}_1, \mathbf{x}_2, \dots, \mathbf{x}_N]$, that are drawn from a union of $L$ low-dimensional subspaces $\{ \cS_\ell \}_{\ell=1}^L$ of dimensions $\{ d_\ell \}_{\ell=1}^L$. The task of subspace clustering is to segment the data points according to their respective subspaces. Low-rank representation (LRR) is a recently proposed subspace clustering method \cite{LiuLY.ICML2010,LiuLYSYM.PAMI2013} that aims to find the lowest-rank representation for the data using a predefined dictionary to reveal the intrinsic geometric structure of the data. Mathematically, LRR can be formulated as the following optimization problem \cite{LiuLY.ICML2010}:
\begin{align}
\widehat{\mathbf{Z}} = \argmin_{\mathbf{Z}} \rank(\mathbf{Z}) \quad \text{s.t.} \quad \mathbf{X} = \mathbf{A}\mathbf{Z},
\end{align}
where $\mathbf{A}$ is a predefined dictionary that linearly spans the data space and $\mathbf{Z}$ is the low-rank representation of the data over $\mathbf{A}$. In practice, the observed data $\mathbf{X}$ are often corrupted by noise. To better handle the noisy data, the LRR problem can be formulated as
\begin{align}    \label{eqn:lrr}
(\widehat{\mathbf{Z}},\widehat{\mathbf{E}}) = \argmin_{\mathbf{Z},\mathbf{E}} \|\mathbf{Z}\|_{\ast} + \lambda \|\mathbf{E}\|_{\iota} \quad \text{s.t.} \quad \mathbf{X} = \mathbf{A}\mathbf{Z} + \mathbf{E},
\end{align}
where $\mathbf{E}$ is a matrix representing the approximation errors of the data, the nuclear norm is the convex relaxation of rank operator and $\|\cdot\|_{\iota}$ indicates a certain regularization strategy involving $\mathbf{E}$. In \cite{LiuLY.ICML2010}, the $\ell_{2,1}$ norm is used for regularization because of its robustness against corruptions and outliers in data. Finally, $\lambda$ is a positive parameter that sets the tradeoff between low rankness of $\mathbf{Z}$ and the representation fidelity. In \cite{LiuLY.ICML2010}, the whole sample set $\mathbf{X}$ is used as the dictionary for clustering, which takes advantage of the self-expressiveness property of data. Once the matrix $\widehat{\mathbf{Z}}$ is available, a symmetric non-negative similarity matrix $\mathbf{W}$ can be defined as $\mathbf{W} = \frac{|\widehat{\mathbf{Z}}| + |\widehat{\mathbf{Z}}^T|}{2}$, where $|\cdot|$ denotes the element-wise absolute value operation. Finally, spectral clustering \cite{NgJW.NIPS2001} can be performed on $\mathbf{W}$ to get the final clustering results.

It has been proved that LRR can achieve a block-diagonal (up to permutation) solution under the condition that the subspaces underlying the data are independent \cite{LiuLY.ICML2010,LiuLYSYM.PAMI2013}. However, clustering of disjoint subspaces is more desirable in many real situations \cite{ElhamifarV.PAMI2013}.\footnote{Heuristically, a collection of $L$ subspaces $\{ \cS_\ell \}_{\ell=1}^L$ is said to be independent if the bases of all the subspaces are linearly independent, whereas $\{ \cS_\ell \}_{\ell=1}^L$ are said to be disjoint if every pair of subspaces are independent; we refer the reader to \cite{ElhamifarV.PAMI2013} for formal definitions.} To improve upon LRR for disjoint subspace clustering, Tang \emph{et al.} \cite{TangLSZ.TNNLS2014} proposed structure-constrained LRR (SC-LRR) model, whose learning can be formulated as follows:
\begin{align}    \label{eqn:sclrr}
(\widehat{\mathbf{Z}},\widehat{\mathbf{E}}) = \argmin_{\mathbf{Z},\mathbf{E}} \|\mathbf{Z}\|_{\ast} + \alpha \|\mathbf{B} \odot \mathbf{Z}\|_1 + \lambda \|\mathbf{E}\|_{2,1} \quad \text{s.t.} \quad \mathbf{X} = \mathbf{X}\mathbf{Z} + \mathbf{E},
\end{align}
where $\alpha$ and $\lambda$ are penalty parameters, $\mathbf{B} \in \R^{N \times N}$ is a predefined weight matrix associated with the data, and $\odot$ denotes the Hadamard product. It has been shown in \cite{TangLSZ.TNNLS2014} that by designing some predefined weight matrices, the optimal solution of $\mathbf{Z}$ is block-diagonal for disjoint subspaces when the data are noiseless. In general, the matrix $\mathbf{B}$ imposes restrictions on the solution by penalizing affinities between data samples from different clusters, while rewarding affinities between data samples from the same cluster. The sample set $\mathbf{X}$ is again selected as the dictionary in \cite{TangLSZ.TNNLS2014} for clustering.

\subsection{CS-LRR Model}
\label{ssec:model}

Almost all the existing subspace clustering methods follow a two-stage approach: ($i$) learning the coefficient matrix from the data and ($ii$) applying spectral clustering on the affinity matrix to segment the data. This two-step approach may lead to sub-optimal clustering results because the final clustering result is independent of the optimization problem that is used to obtain the coefficient matrix. We hypothesize that by making the final clustering result dependent on the generation of the optimal coefficient matrix, we will be able to obtain better clustering results. Specifically, suppose we have the coefficient matrix $\widehat{\mathbf{Z}}$ for SC-LRR. Then one can define an affinity matrix $\mathbf{W}$ as $\mathbf{W} = \frac{|\widehat{\mathbf{Z}}| + |\widehat{\mathbf{Z}}^T|}{2}$. We obtain the clustering of the data by applying spectral clustering \cite{NgJW.NIPS2001} on $\mathbf{W}$, which solves the following problem:
\begin{align}    \label{eqn:spectral}
\widehat{\mathbf{F}} = \argmin_{\mathbf{F}} \tr(\mathbf{F}^T (\mathbf{V} - \mathbf{W}) \mathbf{F}) \quad \text{s.t.} \quad \mathbf{F}^T\mathbf{F} = \mathbf{I},
\end{align}
where $\mathbf{F} \in \R^{N \times L}$ is a binary matrix indicating the cluster membership of the data points, i.e., $\forall i$, $f_{i,\ell} = 1$ if $\mathbf{x}_i$ lies in subspace $\cS_\ell$ and $f_{i,\ell} = 0$ otherwise. Here, $\mathbf{V} \in \R^{N \times N}$ is a diagonal matrix with its diagonal elements defined as $v_{i,i} = \sum_j w_{i,j}$. The solution of $\mathbf{F}$ consists of the eigenvectors of the Laplacian matrix $\mathbf{M} = \mathbf{V} - \mathbf{W}$ associated with its smallest $L$ eigenvalues. Note that the objective function of \eqref{eqn:spectral} can also be written as
\begin{align}    \label{eqn:relation}
\tr(\mathbf{F}^T (\mathbf{V} - \mathbf{W}) \mathbf{F}) = \frac{1}{2} \sum_{i,j} w_{i,j} \|\mathbf{f}^i - \mathbf{f}^j\|_2^2 = \sum_{i,j} |z_{i,j}| (\frac{1}{2} \|\mathbf{f}^i - \mathbf{f}^j\|_2^2) = \|\mathbf{\Theta} \odot \mathbf{Z}\|_1,
\end{align}
where $\theta_{i,j} = \frac{1}{2} \|\mathbf{f}^i - \mathbf{f}^j\|_2^2$. In order to capture the relation between $\mathbf{Z}$ and $\mathbf{F}$, we imagine that the exact segmentation matrix $\mathbf{F}$ is known. It can be observed that if $\mathbf{x}_i$ and $\mathbf{x}_j$ lie in different subspaces, i.e., $\mathbf{f}^i \neq \mathbf{f}^j$, then we would like to have $z_{i,j} = 0$ for better clustering. Therefore, we can use \eqref{eqn:relation} to quantify the disagreement between $\mathbf{Z}$ and $\mathbf{F}$ \cite{LiV.CVPR2015}.

The ground truth segmentation matrix $\mathbf{F}$ is of course unknown in practice. In order to penalize the ``disagreement'' between $\mathbf{Z}$ and $\mathbf{F}$, we propose a \emph{clustering-aware structure-constrained LRR} (CS-LRR) model obtained by solving the following problem:
\begin{align}    \label{eqn:probform}
& \min_{\mathbf{Z},\mathbf{F},\mathbf{E}} \|\mathbf{Z}\|_{\ast} + \alpha \|\mathbf{B} \odot \mathbf{Z}\|_1 + \beta \tr \Big(\mathbf{F}^T (\mathbf{V} - \frac{|\mathbf{Z}| + |\mathbf{Z}^T|}{2}) \mathbf{F}\Big) + \lambda \|\mathbf{E}\|_{\iota}    \nonumber \\
&\quad \text{s.t.} \quad \mathbf{X} = \mathbf{X}\mathbf{Z} + \mathbf{E}, ~ \mathbf{F}^T\mathbf{F} = \mathbf{I},
\end{align}
where $\alpha$, $\beta$ and $\lambda$ are penalty parameters. Similar to \cite{TangLSZ.TNNLS2014}, the $(i,j)$-th entry of $\mathbf{B}$ is defined as $b_{i,j} = 1 - \exp(-\frac{1 - |\mathbf{x}_i^T \mathbf{x}_j|}{\sigma})$, where $\sigma$ is the mean of all $1 - |\mathbf{x}_i^T \mathbf{x}_j|$'s. The CS-LRR model in \eqref{eqn:probform} requires knowledge of the number of subspaces $L$. Initially, we assume to have knowledge of an upper bound on $L$, which we denote by $L_{\mathrm{max}}$, and we use $L_{\mathrm{max}}$ in \eqref{eqn:probform} to learn the representation matrix. In practice, however, one cannot assume knowledge of this parameter a priori. Therefore, we also develop a hierarchical clustering technique to automatically determine $L_{\mathrm{max}}$, which will be discussed in Section~\ref{sec:hcslrr}. The CS-LRR model encourages consistency between the representation coefficients and the subspace segmentation by making the similarity matrix more block-diagonal, which can help spectral clustering achieve the best results.

\subsection{Solving CS-LRR}
\label{ssec:solution}

To solve the optimization problem \eqref{eqn:probform}, we first introduce an auxiliary variable $\mathbf{Q}$ to make the objective function of \eqref{eqn:probform} separable and reformulate \eqref{eqn:probform} as
\begin{align}    \label{eqn:probconstrain}
& \min_{\mathbf{Z},\mathbf{F},\mathbf{E}} \|\mathbf{Z}\|_{\ast} + \alpha \|\mathbf{B} \odot \mathbf{Q}\|_1 + \beta \tr \Big(\mathbf{F}^T (\mathbf{V} - \frac{|\mathbf{Q}| + |\mathbf{Q}^T|}{2}) \mathbf{F}\Big) + \lambda \|\mathbf{E}\|_{\iota}    \nonumber \\
&\quad \text{s.t.} \quad \mathbf{X} = \mathbf{X}\mathbf{Z} + \mathbf{E}, ~ \mathbf{F}^T\mathbf{F} = \mathbf{I}, ~ \mathbf{Z} = \mathbf{Q}.
\end{align}
This problem can be solved by using the linearized alternating direction method (LADM) \cite{LinLS.NIPS2011}. Specifically, the augmented Lagrangian function of \eqref{eqn:probconstrain} is
\begin{align}    \label{eqn:probaug}
& \cL(\mathbf{Z},\mathbf{Q},\mathbf{F},\mathbf{E},\mathbf{\Gamma}_1,\mathbf{\Gamma}_2)    \nonumber \\
& = \|\mathbf{Z}\|_{\ast} + \alpha \|\mathbf{B} \odot \mathbf{Q}\|_1 + \beta \tr \Big(\mathbf{F}^T (\mathbf{V} - \frac{|\mathbf{Q}| + |\mathbf{Q}^T|}{2}) \mathbf{F}\Big) + \lambda \|\mathbf{E}\|_{\iota}    \nonumber \\
&\quad + \langle \mathbf{\Gamma}_1, \mathbf{X} - \mathbf{X}\mathbf{Z} - \mathbf{E} \rangle + \langle \mathbf{\Gamma}_2, \mathbf{Z} - \mathbf{Q} \rangle + \frac{\mu}{2} (\|\mathbf{X} - \mathbf{X}\mathbf{Z} - \mathbf{E}\|_F^2 + \|\mathbf{Z} - \mathbf{Q}\|_F^2),
\end{align}
where $\mathbf{\Gamma}_1$ and $\mathbf{\Gamma}_2$ are matrices of Lagrangian multipliers and $\mu$ is a penalty parameter. The optimization of \eqref{eqn:probaug} can be done iteratively by minimizing $\cL$ with respect to $\mathbf{Z}$, $\mathbf{Q}$, $\mathbf{F}$ and $\mathbf{E}$ one at a time, with all other variables being fixed. Note that we also update $\mathbf{V}$ accordingly once we have $\mathbf{Q}$ updated. The constraint $\mathbf{F}^T\mathbf{F} = \mathbf{I}$ in \eqref{eqn:probconstrain} is imposed independently in each step of updating $\mathbf{F}$.

\emph{Update $\mathbf{Z}$ while fixing other variables:} When other variables are fixed, the problem of updating $\mathbf{Z}$ in the $(t+1)$-th iteration ($t \geq 0$) is equivalent to minimizing the following function:
\begin{align}    \label{eqn:probZ}
f(\mathbf{Z}) = \|\mathbf{Z}\|_{\ast} + \hbar(\mathbf{Z},\mathbf{Q}^t,\mathbf{E}^t,\mathbf{\Gamma}_1^t,\mathbf{\Gamma}_2^t),
\end{align}
where $\hbar(\mathbf{Z},\mathbf{Q}^t,\mathbf{E}^t,\mathbf{\Gamma}_1^t,\mathbf{\Gamma}_2^t) = \frac{\mu^t}{2} (\|\mathbf{X} - \mathbf{X}\mathbf{Z} - \mathbf{E}^t + \frac{\mathbf{\Gamma}_1^t}{\mu^t}\|_F^2 + \|\mathbf{Z} - \mathbf{Q}^t + \frac{\mathbf{\Gamma}_2^t}{\mu^t}\|_F^2)$. However, this variant of the problem does not have a closed-form solution. Nonetheless, in the spirit of LADM, $f(\mathbf{Z})$ can also be minimized by solving the following problem:
\begin{align}    \label{eqn:reformprobZ}
\mathbf{Z}^{t+1} & = \argmin_{\mathbf{Z}} \|\mathbf{Z}\|_{\ast} + \langle \nabla_{\mathbf{Z}} \hbar(\mathbf{Z}^t), \mathbf{Z} - \mathbf{Z}^t \rangle + \frac{\eta \mu^t}{2} \|\mathbf{Z} - \mathbf{Z}^t\|_F^2    \nonumber \\
& = \argmin_{\mathbf{Z}} \|\mathbf{Z}\|_{\ast} + \frac{\eta \mu^t}{2} \|\mathbf{Z} - \mathbf{Z}^t + \frac{\nabla_{\mathbf{Z}} \hbar(\mathbf{Z}^t)}{\eta \mu^t}\|_F^2,
\end{align}
where $\nabla_{\mathbf{Z}} \hbar$ is the partial differential of $\hbar$ with respect to $\mathbf{Z}$ and $\eta$ is a constant satisfying $\eta > \|\mathbf{X}\|^2$. For this problem, $\nabla_{\mathbf{Z}} \hbar = \mu^t \mathbf{X}^T (\mathbf{X}\mathbf{Z} - \mathbf{X} + \mathbf{E}^t - \frac{\mathbf{\Gamma}_1^t}{\mu^t}) + \mu^t (\mathbf{Z} - \mathbf{Q}^t + \frac{\mathbf{\Gamma}_2^t}{\mu^t})$. Then the closed-form solution for $\mathbf{Z}$ is given as
\begin{align}    \label{eqn:solZ}
\mathbf{Z}^{t+1} = \Upsilon_{\frac{1}{\eta \mu^t}} (\mathbf{Z}^t - \frac{\nabla_{\mathbf{Z}} \hbar(\mathbf{Z}^t)}{\eta \mu^t}),
\end{align}
where $\Upsilon (\cdot)$ denotes singular value thresholding operator \cite{CaiCS.SIAM2010}.

\emph{Update $\mathbf{Q}$ while fixing other variables:} When other variables are fixed, the problem of updating $\mathbf{Q}$ is
\begin{align}    \label{eqn:probQ}
\min_{\mathbf{Q}} \alpha \|\mathbf{B} \odot \mathbf{Q}\|_1 + \beta \tr \Big( (\mathbf{F}^t)^T (\mathbf{V} - \frac{|\mathbf{Q}| + |\mathbf{Q}^T|}{2}) \mathbf{F}^t \Big) + \frac{\mu^t}{2} \|\mathbf{Q} - (\mathbf{Z}^{t+1} + \frac{\mathbf{\Gamma}_2^t}{\mu^t})\|_F^2.
\end{align}
According to \eqref{eqn:relation}, we have $\tr ( (\mathbf{F}^t)^T (\mathbf{V} - \frac{|\mathbf{Q}| + |\mathbf{Q}^T|}{2}) \mathbf{F}^t ) = \|\mathbf{\Theta}^t \odot \mathbf{Q}\|_1$, where $\mathbf{\theta}_{i,j}^t = \frac{1}{2} \|\mathbf{f}^{t,i} - \mathbf{f}^{t,j}\|_2^2$. Here, $\mathbf{f}^{t,i}$ and $\mathbf{f}^{t,j}$ denote the $i$-th and $j$-th row of the matrix $\mathbf{F}^t$, respectively. Therefore, \eqref{eqn:probQ} can be written as follows:
\begin{align*}
\min_{\mathbf{Q}} \alpha \|(\mathbf{B} + \frac{\beta}{\alpha} \mathbf{\Theta}^t) \odot \mathbf{Q}\|_1 + \frac{\mu^t}{2} \|\mathbf{Q} - (\mathbf{Z}^{t+1} + \frac{\mathbf{\Gamma}_2^t}{\mu^t})\|_F^2,
\end{align*}
which has the following closed-form solution:
\begin{align}    \label{eqn:solQ}
\mathbf{Q}^{t+1} = \cT_{(\frac{\alpha}{\mu^t},\mathbf{B} + \frac{\beta}{\alpha} \mathbf{\Theta}^t)} (\mathbf{Z}^{t+1} + \frac{\mathbf{\Gamma}_2^t}{\mu^t}),
\end{align}
where the $(i,j)$-th entry of $\cT_{(\tau,\mathbf{H})} (\mathbf{A})$ is given by $T_{\tau h_{i,j}} (a_{i,j})$ with $T_{\tau'}(x) = \max(x-\tau',0) + \min(x+\tau',0)$ \cite{TangLSZ.TNNLS2014}. After this, we update $\mathbf{V}^{t+1}$ by setting $\forall i$, $v_{i,i}^{t+1} = \sum_j \frac{|q_{i,j}^{t+1}| + |q_{j,i}^{t+1}|}{2}$.

\emph{Update $\mathbf{F}$ while fixing other variables:} When other variables are fixed, the problem of updating $\mathbf{F}$ is
\begin{align*}
\mathbf{F}^{t+1} = \argmin_{\mathbf{F}^T\mathbf{F} = \mathbf{I}} \tr(\mathbf{F}^T (\mathbf{V}^{t+1} - \frac{|\mathbf{Q}^{t+1}| + |(\mathbf{Q}^{t+1})^T|}{2}) \mathbf{F}).
\end{align*}
Defining $\mathbf{M}^{t+1} = \mathbf{V}^{t+1} - \frac{|\mathbf{Q}^{t+1}| + |(\mathbf{Q}^{t+1})^T|}{2}$, this problem has a closed-form solution that involves eigendecomposition of $\mathbf{M}^{t+1}$. In particular, the columns of $\mathbf{F}^{t+1}$ are given by the eigenvectors of $\mathbf{M}^{t+1}$ associated with its smallest $L_{\mathrm{max}}$ eigenvalues.

\emph{Update $\mathbf{E}$ while fixing other variables:} When other variables are fixed, the problem of updating $\mathbf{E}$ can be written as
\begin{align}    \label{eqn:probE}
\mathbf{E}^{t+1} = \argmin_{\mathbf{E}} \lambda \|\mathbf{E}\|_{\iota} + \frac{\mu^t}{2} \|\mathbf{E} - \mathbf{C}^{t+1}\|_F^2,
\end{align}
where $\mathbf{C}^{t+1} = \mathbf{X} - \mathbf{X}\mathbf{Z}^{t+1} + \frac{\mathbf{\Gamma}_1^t}{\mu^t}$. For HOG features, we define $\widehat{\mathbf{E}}_i$ to be the approximation error with respect to $\widehat{\mathbf{X}}_i$ (the matrix version of $\mathbf{x}_i$) and set the error term $\|\mathbf{E}\|_{\iota} \equiv \sum_{i=1}^N \|\widehat{\mathbf{E}}_i\|_{2,1}$ to ensure robustness against ``corruptions'' in the orientation of each HOG feature descriptor; this is because the background information is included in the feature vector. Then \eqref{eqn:probE} can be decomposed into $N$ independent subproblems. In order to update $\mathbf{e}_i$, we first convert the vector $\mathbf{c}_i^{t+1}$ to a matrix $\widehat{\mathbf{C}}_i^{t+1} \in \R^{n_b \times 9}$ and then solve the following problem:
\begin{align}
\widehat{\mathbf{E}}_i^{t+1} = \argmin_{\widehat{\mathbf{E}}_i} \lambda \|\widehat{\mathbf{E}}_i\|_{2,1} + \frac{\mu^t}{2} \|\widehat{\mathbf{E}}_i - \widehat{\mathbf{C}}_i^{t+1}\|_F^2,
\end{align}
where $\widehat{\mathbf{E}}_i^{t+1} \in \R^{n_b \times 9}$ is the reshaped ``image'' of the vector $\mathbf{e}_i^{t+1}$. This problem can be solved using \cite[Lemma 3.2]{LiuLY.ICML2010}. For MBH features, since the noise due to background motion is eliminated, we simply set the error term $\|\mathbf{E}\|_{\iota} \equiv \|\mathbf{E}\|_{2,1}$; then \eqref{eqn:probE} can be written as
\begin{align}
\mathbf{E}^{t+1} = \argmin_{\mathbf{E}} \lambda \|\mathbf{E}\|_{2,1} + \frac{\mu^t}{2} \|\mathbf{E} - \mathbf{C}^{t+1}\|_F^2,
\end{align}
which can also be solved by using \cite[Lemma 3.2]{LiuLY.ICML2010}. The complete algorithm is outlined in Algorithm~\ref{algo:cslrr}.

\algsetup{indent=0.2em}
\begin{algorithm}[t]
\caption{Solving CS-LRR by LADM}
\label{algo:cslrr}
\textbf{Input:} The data matrix $\mathbf{X}$ and matrix $\mathbf{B}$, and parameters $L_{\mathrm{max}}$, $\alpha$, $\beta$ and $\lambda$. \\
\textbf{Initialize:} $\mathbf{Z}^0=\mathbf{Q}^0=\mathbf{\Theta}^0=\mathbf{\Gamma}_1^0=\mathbf{\Gamma}_2^0=\mathbf{0}$, $\rho=1.1$, $\eta > \|\mathbf{X}\|^2$, $\mu^0=0.1$, $\mu_{\mathrm{max}}=10^{30}$, $\epsilon=10^{-7}$, $t=0$.

\begin{algorithmic}[1]
\WHILE{not converged}
\STATE Fix other variables and update $\mathbf{Z}$: \\
$\mathbf{Z}^{t+1} = \argmin_{\mathbf{Z}} \|\mathbf{Z}\|_{\ast} + \frac{\eta \mu^t}{2} \|\mathbf{Z} - \mathbf{Z}^t + (\mathbf{X}^T (\mathbf{X}\mathbf{Z} - \mathbf{X} + \mathbf{E}^t - \frac{\mathbf{\Gamma}_1^t}{\mu^t}) + (\mathbf{Z} - \mathbf{Q}^t + \frac{\mathbf{\Gamma}_2^t}{\mu^t})) /\eta\|_F^2$.
\STATE Fix other variables and update $\mathbf{Q}$: \\
$\mathbf{Q}^{t+1} = \argmin_{\mathbf{Q}} \alpha \|(\mathbf{B} + \frac{\beta}{\alpha} \mathbf{\Theta}^t) \odot \mathbf{Q}\|_1 + \frac{\mu^t}{2} \|\mathbf{Q} - (\mathbf{Z}^{t+1} + \frac{\mathbf{\Gamma}_2^t}{\mu^t})\|_F^2$.
\STATE Compute the Laplacian matrix: $\mathbf{M}^{t+1} = \mathbf{V}^{t+1} - \frac{|\mathbf{Q}^{t+1}| + |(\mathbf{Q}^{t+1})^T|}{2}$ and update $\mathbf{F}$ by $\mathbf{F}^{t+1} = \argmin_{\mathbf{F}^T\mathbf{F} = \mathbf{I}} \tr(\mathbf{F}^T \mathbf{M}^{t+1} \mathbf{F})$.
\STATE Fix other variables and update $\mathbf{E}$: \\
$\mathbf{E}^{t+1} = \argmin_{\mathbf{E}} \lambda \|\mathbf{E}\|_{\iota} + \frac{\mu^t}{2} \|\mathbf{E} - (\mathbf{X} - \mathbf{X}\mathbf{Z}^{t+1} + \frac{\mathbf{\Gamma}_1^t}{\mu^t})\|_F^2$.
\STATE Update Lagrange multipliers: \\
$\mathbf{\Gamma}_1^{t+1} = \mathbf{\Gamma}_1^t + \mu^t (\mathbf{X} - \mathbf{X}\mathbf{Z}^{t+1} - \mathbf{E}^{t+1})$, \\
$\mathbf{\Gamma}_2^{t+1} = \mathbf{\Gamma}_2^t + \mu^t (\mathbf{Z}^{t+1} - \mathbf{Q}^{t+1})$.
\STATE Update $\mu^{t+1}$ as $\mu^{t+1} = \min(\mu_{\mathrm{max}},\rho \mu^t)$.
\STATE Check convergence conditions and break if \\
$\|\mathbf{X} - \mathbf{X}\mathbf{Z}^{t+1} - \mathbf{E}^{t+1}\|_{\infty} \leq \epsilon$, $\|\mathbf{Z}^{t+1} - \mathbf{Q}^{t+1}\|_{\infty} \leq \epsilon$.
\STATE Update $t$ by $t = t + 1$.
\ENDWHILE
\end{algorithmic}

\textbf{Output:} The optimal low-rank representation $\widehat{\mathbf{Z}} = \mathbf{Z}^t$.
\end{algorithm}

\section{Hierarchical Subspace Clustering Based on CS-LRR Model}
\label{sec:hcslrr}

We now introduce a hierarchical subspace clustering algorithm based on CS-LRR approach for learning action attributes at multiple levels of our UoS model and for automatically determining the final number of attributes present in a high-dimensional dataset without prior knowledge. To begin, we introduce some notation used in this section. We define $\boldsymbol{\pi}_{p|\ell}$ to be the set containing the indexes of all $\mathbf{x}_i$'s that are assigned to the $\ell$-th subspace at the $p$-th level ($p \geq 1$) of the hierarchical structure, and let $\mathbf{X}_{p|\ell} = [\mathbf{x}_i: i \in \boldsymbol{\pi}_{p|\ell}] \in \R^{m \times N_{p|\ell}}$ be the corresponding set of signals, where $N_{p|\ell}$ is the number of signals in $\mathbf{X}_{p|\ell}$. Let $L_p$ denote the number of subspaces at the $p$-th level, then we have $\sum_{\ell=1}^{L_p} N_{p|\ell} = N$ and $\mathbf{X} = \bigcup_{\ell=1}^{L_p} \mathbf{X}_{p|\ell}$ for all $p$'s. The subspace underlying $\mathbf{X}_{p|\ell}$ is denoted by $\cS_{p|\ell}$ and the dimension of the subspace $\cS_{p|\ell}$ is denoted by $d_{p|\ell}$.

We first apply Algorithm~\ref{algo:cslrr} to obtain the optimal representation coefficient matrix $\widehat{\mathbf{Z}}$. Then we set the coefficients below a given threshold to zeros, and we denote the final representation matrix by $\widetilde{\mathbf{Z}}$. By defining the affinity matrix $\mathbf{W} = \frac{|\widetilde{\mathbf{Z}}| + |\widetilde{\mathbf{Z}}^T|}{2}$, we proceed with our hierarchical clustering procedure as follows. We begin by applying spectral clustering \cite{NgJW.NIPS2001} based on $\mathbf{W}$ at the first level ($p = 1$), which divides $\mathbf{X}$ into two subspaces with clusters $\mathbf{X} = \mathbf{X}_{1|1} \cup \mathbf{X}_{1|2}$ such that $\mathbf{X}_{1|1} \cap \mathbf{X}_{1|2} = \emptyset$, and we use $\boldsymbol{\pi}_{1|\ell}$'s ($\ell = 1, 2$) to denote the indexes of signals in each cluster. At the second level, we perform spectral clustering based on $[\mathbf{W}]_{\boldsymbol{\pi}_{1|1},\boldsymbol{\pi}_{1|1}}$ and $[\mathbf{W}]_{\boldsymbol{\pi}_{1|2},\boldsymbol{\pi}_{1|2}}$ separately and divide each $\mathbf{X}_{1|\ell}$ ($\ell = 1, 2$) into 2 clusters, yielding 4 clusters $\mathbf{X} = \bigcup_{\ell=1}^{4} \mathbf{X}_{2|\ell}$ with $\mathbf{X}_{1|\ell} = \mathbf{X}_{2|2\ell-1} \cup \mathbf{X}_{2|2\ell}$ ($\ell = 1, 2$). Using the signals in $\mathbf{X}_{2|\ell}$ ($\ell = 1, \dots, 4$), we estimate the four subspaces $\cS_{2|\ell}$'s underlying $\mathbf{X}_{2|\ell}$'s by identifying their orthonormal bases. To be specific, we obtain eigendecomposition of the covariance matrix $\mathbf{Y}_{2|\ell} = \mathbf{X}_{2|\ell} \mathbf{X}_{2|\ell}^T$ such that $\mathbf{Y}_{2|\ell} = \mathbf{U}_{2|\ell} \mathbf{\Lambda}_{2|\ell} \mathbf{U}_{2|\ell}^T$, where $\mathbf{\Lambda}_{2|\ell} = \diag (\lambda_1, \dots, \lambda_{N_{2|\ell}})$ is a diagonal matrix ($\lambda_1 \geq \lambda_2 \geq \dots \geq \lambda_{N_{2|\ell}}$) and $\mathbf{U}_{2|\ell} = [\mathbf{u}_1, \dots, \mathbf{u}_{N_{2|\ell}}]$. Then the dimension of the subspace $\cS_{2|\ell}$, denoted by $d_{2|\ell}$, is estimated based on the energy threshold, i.e., $d_{2|\ell} = \argmin_{d} \frac{\sum_{\omega=1}^{d} \lambda_\omega}{\sum_{\omega=1}^{N_{2|\ell}} \lambda_\omega} \geq \gamma$, where $\gamma$ is a predefined threshold and is set close to 1 for better representation. The orthonormal basis of $\cS_{2|\ell}$ can then be written as $\widehat{\mathbf{U}}_{2|\ell} = [\mathbf{u}_1, \dots, \mathbf{u}_{d_{2|\ell}}]$. After this step, we end up with 4 clusters $\{ \mathbf{X}_{2|\ell} \}_{\ell=1}^{4}$ with their corresponding indexes $\{ \boldsymbol{\pi}_{2|\ell} \}_{\ell=1}^{4}$ and associated orthonormal bases $\{ \widehat{\mathbf{U}}_{2|\ell} \}_{\ell=1}^{4}$.

\algsetup{indent=0.2em}
\begin{algorithm}[t]
\caption{HCS-LRR (the $p$-th level, $2 \leq p \leq P-1$)}
\label{algo:hcslrr}
\textbf{Input:} The affinity matrix $\mathbf{W}$ obtained from Algorithm~\ref{algo:cslrr}. A set of clusters $\{ \mathbf{X}_{p|\ell} \}_{\ell=1}^{L_p}$ with their corresponding indexes $\{ \boldsymbol{\pi}_{p|\ell} \}_{\ell=1}^{L_p}$, their underlying subspace bases $\{ \widehat{\mathbf{U}}_{p|\ell} \}_{\ell=1}^{L_p}$ and $\{ g_{p|\ell} \}_{\ell=1}^{L_p}$, and parameters $\gamma$, $\varrho$ and $d_{\mathrm{min}}$.

\begin{algorithmic}[1]
\STATE $\widehat{L} = 0$.
\FORALL{$\ell = 1$ to $L_p$}
\IF{$g_{p|\ell} = 1$}
\STATE $\theta = 0$.
\STATE Apply spectral clustering on $[\mathbf{W}]_{\boldsymbol{\pi}_{p|\ell},\boldsymbol{\pi}_{p|\ell}}$ to split $\mathbf{X}_{p|\ell}$ into $\mathbf{X}_{p|\ell} = \mathbf{\Sigma}_1 \cup \mathbf{\Sigma}_2$ ($\boldsymbol{\pi}_{p|\ell} = \boldsymbol{\chi}_1 \cup \boldsymbol{\chi}_2$).
\STATE $\forall c = 1, 2$, estimate $d_{\mathbf{\Sigma}_c}$ and $\widehat{\mathbf{U}}_{\mathbf{\Sigma}_c}$ of $\cS_{\mathbf{\Sigma}_c}$ using $\mathbf{\Sigma}_c$.
\STATE $\forall c = 1, 2$, compute $\bar{\delta}_c$ and $\bar{\zeta}_c$. If $\frac{\bar{\delta}_c - \bar{\zeta}_c}{\bar{\delta}_c} \geq \varrho$, $\theta = \theta + 1$.
\IF{$\theta \geq 1$ and $\min(d_{\mathbf{\Sigma}_1},d_{\mathbf{\Sigma}_2}) \geq d_{\mathrm{min}}$}
\STATE $\forall c = 1, 2$, $\mathbf{X}_{p+1|\widehat{L}+c} = \mathbf{\Sigma}_c$, $\boldsymbol{\pi}_{p+1|\widehat{L}+c} = \boldsymbol{\chi}_c$, $\widehat{\mathbf{U}}_{p+1|\widehat{L}+c} = \widehat{\mathbf{U}}_{\mathbf{\Sigma}_c}$, and $g_{p+1|\widehat{L}+c} = 1$.
\STATE $\widehat{L} = \widehat{L} + 2$.
\ELSE
\STATE $\mathbf{X}_{p+1|\widehat{L}+1} = \mathbf{X}_{p|\ell}$, $\boldsymbol{\pi}_{p+1|\widehat{L}+1} = \boldsymbol{\pi}_{p|\ell}$, $\widehat{\mathbf{U}}_{p+1|\widehat{L}+1} = \widehat{\mathbf{U}}_{p|\ell}$, $g_{p|\ell} = 0$, $g_{p+1|\widehat{L}+1} = 0$, and $\widehat{L} = \widehat{L} + 1$.
\ENDIF
\ELSE
\STATE $\mathbf{X}_{p+1|\widehat{L}+1} = \mathbf{X}_{p|\ell}$, $\boldsymbol{\pi}_{p+1|\widehat{L}+1} = \boldsymbol{\pi}_{p|\ell}$, $\widehat{\mathbf{U}}_{p+1|\widehat{L}+1} = \widehat{\mathbf{U}}_{p|\ell}$, $g_{p+1|\widehat{L}+1} = 0$, and $\widehat{L} = \widehat{L} + 1$.
\ENDIF
\ENDFOR
\STATE $L_{p+1} = \widehat{L}$.
\end{algorithmic}

\textbf{Output:} A set of clusters $\{ \mathbf{X}_{p+1|\ell} \}_{\ell=1}^{L_{p+1}}$ with their corresponding indexes $\{ \boldsymbol{\pi}_{p+1|\ell} \}_{\ell=1}^{L_{p+1}}$, orthonormal bases of the attributes $\{ \widehat{\mathbf{U}}_{p+1|\ell} \}_{\ell=1}^{L_{p+1}}$ and $\{ g_{p+1|\ell} \}_{\ell=1}^{L_{p+1}}$.
\end{algorithm}

For every $p \geq 2$, we decide whether or not to further divide each single cluster (i.e., subspace) at the $p$-th level into two clusters (subspaces) at the ($p+1$)-th level based on the following principle. We use a binary variable $g_{p|\ell}$ to indicate whether the cluster $\mathbf{X}_{p|\ell}$ is further divisible at the next level or not. If it is, we set $g_{p|\ell} = 1$, otherwise $g_{p|\ell} = 0$. We initialize $g_{2|\ell} = 1$ for all $\ell$'s ($\ell = 1, \dots, 4$). Consider the cluster $\mathbf{X}_{p|\ell}$ at the $p$-th level and assume there already exist $\widehat{L}$ clusters at the ($p+1$)-th level derived from $\{ \mathbf{X}_{p|1}, \mathbf{X}_{p|2}, \dots, \mathbf{X}_{p|\ell-1} \}$. If $g_{p|\ell} = 0$, the ($\widehat{L}+1$)-th cluster at the ($p+1$)-th level will be the same as $\mathbf{X}_{p|\ell}$; thus, we simply set $\mathbf{X}_{p+1|\widehat{L}+1} = \mathbf{X}_{p|\ell}$, $\boldsymbol{\pi}_{p+1|\widehat{L}+1} = \boldsymbol{\pi}_{p|\ell}$, $\widehat{\mathbf{U}}_{p+1|\widehat{L}+1} = \widehat{\mathbf{U}}_{p|\ell}$, and $g_{p+1|\widehat{L}+1} = 0$. If $g_{p|\ell} = 1$, we first split $\mathbf{X}_{p|\ell}$ into two sub-clusters $\mathbf{X}_{p|\ell} = \mathbf{\Sigma}_1 \cup \mathbf{\Sigma}_2$ by applying spectral clustering on $[\mathbf{W}]_{\boldsymbol{\pi}_{p|\ell},\boldsymbol{\pi}_{p|\ell}}$, and we use $\boldsymbol{\chi}_c$ ($c = 1, 2$) to be the set containing the indexes of the signals in $\mathbf{\Sigma}_c$. Then we find the subspaces $\cS_{\mathbf{\Sigma}_c}$ ($c = 1, 2$) underlying $\mathbf{\Sigma}_c$'s respectively using the aforementioned strategy, while their dimensions and orthonormal bases are denoted by $d_{\mathbf{\Sigma}_c}$'s and $\widehat{\mathbf{U}}_{\mathbf{\Sigma}_c}$'s, respectively. After this step, we compute the relative representation error of every signal $\mathbf{x}_i$ in $\mathbf{\Sigma}_c$ ($c = 1, 2$) using the parent subspace basis $\widehat{\mathbf{U}}_{p|\ell}$ and the child subspace basis $\widehat{\mathbf{U}}_{\mathbf{\Sigma}_c}$, which are defined as $\delta_i = \frac{\|\mathbf{x}_i - \widehat{\mathbf{U}}_{p|\ell} \widehat{\mathbf{U}}_{p|\ell}^T \mathbf{x}_i\|_2^2}{\|\mathbf{x}_i\|_2^2}$ and $\zeta_i = \frac{\|\mathbf{x}_i - \widehat{\mathbf{U}}_{\mathbf{\Sigma}_c} \widehat{\mathbf{U}}_{\mathbf{\Sigma}_c}^T \mathbf{x}_i\|_2^2}{\|\mathbf{x}_i\|_2^2}$, respectively. We use $\bar{\delta}_c$ and $\bar{\zeta}_c$ to denote the mean of $\delta_i$'s and $\zeta_i$'s in $\mathbf{\Sigma}_c$ ($c = 1, 2$), respectively. Finally, we say $\mathbf{X}_{p|\ell}$ is divisible if ($i$) the relative representation errors of the signals using the child subspace are less than the representation errors calculated using the parent subspace by a certain threshold, i.e., $\frac{\bar{\delta}_c - \bar{\zeta}_c}{\bar{\delta}_c} \geq \varrho$ for either $c = 1$ or 2, and ($ii$) the dimensions of the two child subspaces meet a minimum requirement, that is, $\min(d_{\mathbf{\Sigma}_1},d_{\mathbf{\Sigma}_2}) \geq d_{\mathrm{min}}$. In here, $\varrho$ and $d_{\mathrm{min}}$ are user-defined parameters and are set to avoid redundant subspaces. When either $\varrho$ or $d_{\mathrm{min}}$ decreases, we tend to have more subspaces at every level $p > 2$. Assuming the two conditions are satisfied, the cluster $\mathbf{X}_{p|\ell}$ is then divided by setting $\mathbf{X}_{p+1|\widehat{L}+1} = \mathbf{\Sigma}_1$ ($\boldsymbol{\pi}_{p+1|\widehat{L}+1} = \boldsymbol{\chi}_1$, $g_{p+1|\widehat{L}+1} = 1$) and $\mathbf{X}_{p+1|\widehat{L}+2} = \mathbf{\Sigma}_2$ ($\boldsymbol{\pi}_{p+1|\widehat{L}+2} = \boldsymbol{\chi}_2$, $g_{p+1|\widehat{L}+2} = 1$). The bases of the subspaces at the ($p+1$)-th level are set by $\widehat{\mathbf{U}}_{p+1|\widehat{L}+1} = \widehat{\mathbf{U}}_{\mathbf{\Sigma}_1}$ and $\widehat{\mathbf{U}}_{p+1|\widehat{L}+2} = \widehat{\mathbf{U}}_{\mathbf{\Sigma}_2}$. If the above conditions are not satisfied, we set $\mathbf{X}_{p+1|\widehat{L}+1} = \mathbf{X}_{p|\ell}$, $\boldsymbol{\pi}_{p+1|\widehat{L}+1} = \boldsymbol{\pi}_{p|\ell}$, $\widehat{\mathbf{U}}_{p+1|\widehat{L}+1} = \widehat{\mathbf{U}}_{p|\ell}$, $g_{p|\ell} = 0$ and $g_{p+1|\widehat{L}+1} = 0$ to indicate $\mathbf{X}_{p|\ell}$, i.e., $\mathbf{X}_{p+1|\widehat{L}+1}$, is a leaf cluster and this cluster will not be divided any further. This process is repeated until we reach a predefined maximum level in the hierarchy denoted by $P$. The hierarchical subspace clustering algorithm based on CS-LRR model for any level $2 \leq p \leq P-1$ is described in Algorithm~\ref{algo:hcslrr}, which we term HCS-LRR. It is worth noting that the maximum number of leaf clusters is $L_{\mathrm{max}} = 2^P$ in this setting, which we set as a key input parameter of Algorithm~\ref{algo:cslrr}.

\section{Attribute Visualization and Semantic Summarization}
\label{sec:semantic}

Given the learned subspaces at different levels of the hierarchical structure, our next goal is to develop a method that helps an expert-in-the-loop to visualize the learned human action attributes, give them semantic labels, and use the labeled attributes to summarize long video sequences of human activities in terms of language at different resolutions. As we have shown previously in \cite{WuGRB.ICCVW2015}, if frame-by-frame silhouette features are used for learning the human action attributes, the attributes (subspaces) can be easily visualized by reshaping the first few vectors of the orthonormal bases of the subspaces into an image format and displaying the scaled versions of these images. However, if other spatial or spatio-temporal features like HOG or MBH are used, the attributes or the subspaces learned using HCS-LRR algorithm cannot be visualized directly by just reshaping each dimension of the subspace in the feature domain.

\subsection{Visualization of Attributes Using HOG Features}
\label{ssec:hogvis}

In the case of HOG features, inspired by HOGgles \cite{VondrickKMT.ICCV2013}, we propose an algorithm to visualize the learned attributes by mapping them back to the pixel domain. In particular, we are interested in building a mapping between the pixel (image) space and the HOG feature space and use this mapping to transform the bases of the HOG feature subspaces into the image space and visualize the attributes. An algorithm based on paired dictionary learning is used to develop this mapping. Concretely, let $\mathbf{X}^I = [\mathbf{x}_1^I, \mathbf{x}_2^I, \dots, \mathbf{x}_{N_p}^I] \in \R^{m_I \times N_p}$ be the collection of $N_p$ vectorized patches of size $m_I = n_{\sigma} \times n_{\sigma}$ pixels from video frames, and $\mathbf{X}^H = [\mathbf{x}_1^H, \mathbf{x}_2^H, \dots, \mathbf{x}_{N_p}^H] \in \R^{m_H \times N_p}$ be the corresponding HOG feature vectors of the patches in $\mathbf{X}^I$. Here, the dimensionality of the HOG features of each patch $m_H$ depends on the choice of the number of bins in the histogram. For better visualization quality, we extract 18-bin contrast specific HOG features. Hence, $m_H = 18$ in this work. Then, two dictionaries, i.e., overcomplete bases whose columns span the data space, $\mathbf{D}_I \in \R^{m_I \times K}$ and $\mathbf{D}_H \in \R^{m_H \times K}$ consisting of $K$ atoms are learned to represent image space and HOG feature space, respectively, such that the sparse representation of any $\mathbf{x}_i^I \in \mathbf{X}^I$ in terms of $\mathbf{D}_I$ should be the same as that of $\mathbf{x}_i^H \in \mathbf{X}^H$ in terms of $\mathbf{D}_H$. Similar to \cite{YangWHM.TIP2010}, paired dictionary learning problem can be considered as solving the following optimization problem:
\begin{align}    \label{eqn:dictlearn}
& \min_{\mathbf{D}_I,\mathbf{D}_H, \{ \boldsymbol{\alpha}_i \}_{i=1}^{N_p}} \sum_{i=1}^{N_p} \Big( \|\mathbf{x}_i^I - \mathbf{D}_I\boldsymbol{\alpha}_i\|_2^2 + \|\mathbf{x}_i^H - \mathbf{D}_H\boldsymbol{\alpha}_i\|_2^2 \Big)    \nonumber \\
&\quad \text{s.t.} \quad \|\boldsymbol{\alpha}_i\|_1 \leq \varepsilon, \forall i, \|\mathbf{d}_{I,k}\|_2 \leq 1, \|\mathbf{d}_{H,k}\|_2 \leq 1, \forall k,
\end{align}
where $\boldsymbol{\alpha}_i$ denotes the sparse code with respect to $\mathbf{x}_i^I$/$\mathbf{x}_i^H$, and $\mathbf{d}_{I,k}$ and $\mathbf{d}_{H,k}$ denote the $k$-th column of $\mathbf{D}_I$ and $\mathbf{D}_H$, respectively.

Equation \eqref{eqn:dictlearn} can be simplified into a standard dictionary learning and sparse coding problem and can be solved using the K-SVD algorithm \cite{AharonEB.TSP2006}. Once we have the collection of HOG feature vectors that lie on a subspace, we can find orthogonal basis of the attribute (subspace) using eigenanalysis. However, these orthogonal basis vectors are not guaranteed to be non-negative, which is the characteristic of HOG feature vectors. Therefore, we use Non-negative Matrix Factorization (NMF) \cite{LeeS.Nature1999} to obtain non-negative basis vectors (not necessarily orthogonal) of each subspace in the feature domain. In order to obtain the corresponding pixel values of each of the basis vectors, we split every basis vector into $n_b$ segments and each segment is of length $m_H$ (since the entire basis vector is the concatenation of HOG features extracted from different patches or blocks in a video frame). After this, we infer the sparse representation of every small segment in the basis vector with respect to $\mathbf{D}_H$ using Orthogonal Matching Pursuit (OMP) \cite{TroppG.TIT2007}, and use the resulting sparse code on $\mathbf{D}_I$ to recover the pixel values corresponding to that patch (which has $m_I$ values) in the frame. This procedure is repeated for all the segments in the basis vector to obtain its image version for visualization. Finally, the subspace can be labeled by visualizing the subspaces, similar to what was done in the case of silhouette features \cite{WuGRB.ICCVW2015}.

\subsection{Visualization of Attributes Using MBH Features}
\label{ssec:mbhvis}

Unlike HOG features, it is not feasible to learn the mapping between the feature domain and the pixel domain for MBH features because most of the patches in the pixel domain will be mapped to zero in the MBH feature domain except the blocks in the action interest region. Instead, we store an intermediate output, i.e., the magnitude of the optical flow for visualization purpose. In other words, for every $n_{\tau}$ consecutive optical flow fields, we use the magnitude of the first optical flow field, which can be visualized by an image, as the representative of all these $n_{\tau}$ optical flow fields. Then for every $n_{\sigma} \times n_{\sigma} \times n_{\tau}$ spatio-temporal patch (block) of optical flow fields, from which MBH features are extracted, we have one $n_{\sigma} \times n_{\sigma}$ optical flow (magnitude) patch associated with this block. We use MBH features for clustering the data samples into different attributes. Then, we use the optical flow frames corresponding to the data samples in each cluster or attribute to visualize the subspace. The optical flow frames look very similar to the silhouette features with non-zero elements present only in the location where there is movement between the original video frames. Thus, the attributes (subspaces) can be easily visualized by reshaping the first few columns of the orthonormal bases of the subspaces, which are now represented by optical flow frames, into an image format.

\section{Action Recognition Using Learned Subspaces}
\label{sec:recogmethod}

In this section, we describe the classification strategy to perform action recognition using the hierarchical union of subspaces learned from Algorithm~\ref{algo:hcslrr}. We assume HCS-LRR ends up with $L_P$ leaf subspaces and the orthonormal bases of these subspaces are being represented by $\{ \widehat{\mathbf{U}}_{P|\ell} \in \R^{m \times d_{P|\ell}} \}_{\ell=1}^{L_P}$. We begin our discussion for the classical multi-class closed set recognition problem.

\subsection{Closed Set Action Recognition Using $k$-NN}
\label{ssec:closedsetkNN}

First, we develop a closed set action recognition method based on a $k$ nearest neighbors classifier. Let $\{ \mathbf{\Phi}_i \in \R^{m \times \xi_i} \}_{i=1}^{N_T}$ be a collection of $N_T$ labeled training samples, where $\xi_i$ denotes the number of feature vectors of the $i$-th training video sequence. Given a test video sequence, whose feature vectors are denoted by $\widehat{\mathbf{\Phi}} = [\widehat{\boldsymbol{\phi}}_1, \widehat{\boldsymbol{\phi}}_2, \dots, \widehat{\boldsymbol{\phi}}_{\widehat{\xi}}]$, we compute the distance between the feature vectors of every training and test sequence as follows. Let $\mathbf{\Phi} = [\boldsymbol{\phi}_1, \boldsymbol{\phi}_2, \dots, \boldsymbol{\phi}_{\xi}]$ be any training sample (we remove subscript $i$ for ease of notation), we first apply Dynamic Time Warping (DTW) \cite{SakoeC.ASSP1978} to align the two action sequences using the HOG/MBH feature vectors and remove redundant segments at the start and the end of the path. We define $\mathbb{P}_{\mathbf{\Phi},\widehat{\mathbf{\Phi}}} = \{ \boldsymbol{\phi}_{a_h}, \widehat{\boldsymbol{\phi}}_{\widehat{a}_h} \}_{h=1}^H$ to be the optimal alignment path with length $H$. Then we assign every vector in these two sequences to the ``closest leaf subspace'' in $\cS_{P|\ell}$'s and we use $\boldsymbol{\psi} \in \R^{\xi}$ and $\widehat{\boldsymbol{\psi}} \in \R^{\widehat{\xi}}$ to denote the vectors which contain the resulting subspace assignment indexes of $\mathbf{\Phi}$ and $\widehat{\mathbf{\Phi}}$, respectively. Based on the optimal alignment path $\mathbb{P}$, the distance $d_F(\mathbf{\Phi},\widehat{\mathbf{\Phi}})$ between these two action sequences is defined as the average of the normalized distances between the leaf subspaces on the alignment path:
\begin{align}    \label{eqn:sequencedist}
d_F(\mathbf{\Phi},\widehat{\mathbf{\Phi}}) = \frac{\sum_{h=1}^{H} d_u(\cS_{P|\psi_{(a_h)}},\cS_{P|\widehat{\psi}_{(\widehat{a}_h)}} )}{H},
\end{align}
where $d_u(\cS_{P|\ell},\cS_{P|\widehat{\ell}}) = \sqrt{ 1 - \frac{ \tr( \widehat{\mathbf{U}}_{P|\ell}^T \widehat{\mathbf{U}}_{P|\widehat{\ell}} \widehat{\mathbf{U}}_{P|\widehat{\ell}}^T \widehat{\mathbf{U}}_{P|\ell} ) } {\max( d_{P|\ell},d_{P|\widehat{\ell}} )} }$ \cite{WangWF.PR2006}. Finally, we use the $k$ nearest neighbors ($k$-NN) classifier to recognize actions based on sequence-to-sequence distances, i.e., a test sequence is declared to be in the class for which the average distance between the test sequence and the $k$ nearest training sequences is the smallest.

\subsection{Closed Set Action Recognition Using SVM}
\label{ssec:closedsetSVM}

Next, we describe a closed set action recognition method based on a non-linear support vector machine (SVM) classifier. Given a collection of $N_T$ labeled training samples, denoted by $\{ \mathbf{\Phi}_i \in \R^{m \times \xi_i} \}_{i=1}^{N_T}$, we first assign every vector in the training samples to the ``closest leaf subspace'' in $\cS_{P|\ell}$'s, and we use $\{ \boldsymbol{\psi}_i \in \R^{\xi_i} \}_{i=1}^{N_T}$ to denote the set of resulting subspace transition vectors. One of the most widely used kernel functions in kernel SVM is Gaussian radial basis function (RBF) kernel $\kappa(\boldsymbol{\psi}_i,\boldsymbol{\psi}_j) = \exp ( - \frac{\|\boldsymbol{\psi}_i - \boldsymbol{\psi}_j\|_2^2}{\nu^2} )$ \cite{Burges.DMKD1998}, where $\nu$ is the bandwidth parameter. However, for action recognition based on the UoS model, the subspace transition vectors of different video sequences have different lengths. Hence, we first use DTW \cite{SakoeC.ASSP1978} on the Grassmann manifold to compute the distance between two action video sequences by only using subspace transition vectors (without aligning the sequences using feature vectors) in Algorithm~\ref{algo:DTW}, which is denoted by $d_T(\cdot,\cdot)$, and replace the Euclidean distance in the Gaussian RBF kernel with DTW distance to obtain a Gaussian DTW kernel \cite{GudmundssonRS.IJCNN2008} as $\kappa(\boldsymbol{\psi}_i,\boldsymbol{\psi}_j) = \exp ( - \frac{ d_T^2(\boldsymbol{\psi}_i,\boldsymbol{\psi}_j) }{\nu^2} )$. Finally, we use both \emph{one-vs.-all} and \emph{one-vs.-one} approach \cite{Burges.DMKD1998} for classification.

\algsetup{indent=0.2em}
\begin{algorithm}[htbp]
\caption{Dynamic Time Warping on the Grassmann Manifold}
\label{algo:DTW}
\textbf{Input:} Two subspace assignment vectors $\boldsymbol{\psi}_i \in \R^{\xi_i}$ and $\boldsymbol{\psi}_j \in \R^{\xi_j}$, leaf subspace bases $\{ \widehat{\mathbf{U}}_{P|\ell} \}_{\ell=1}^{L_P}$ of $\cS_{P|\ell}$'s. \\
\textbf{Initialize:} A matrix $\mathbf{R} \in \R^{(\xi_i+1) \times (\xi_j+1)}$ with $r_{1,1}=0$ and all other entries in the first row and column are $\infty$.

\begin{algorithmic}[1]
\FORALL{$a=1$ to $\xi_i$}
\FORALL{$b=1$ to $\xi_j$}
\STATE $r_{a+1,b+1} = d_u(\cS_{P|{\psi_i}_{(a)}},\cS_{P|{\psi_j}_{(b)}}) + \min ( r_{a,b+1}, r_{a+1,b}, r_{a,b} )$.
\ENDFOR
\ENDFOR
\end{algorithmic}

\textbf{Output:} Distance between $\boldsymbol{\psi}_i$ and $\boldsymbol{\psi}_j$ is $d_T(\boldsymbol{\psi}_i,\boldsymbol{\psi}_j) = r_{\xi_i+1,\xi_j+1}$.
\end{algorithm}

\subsection{Open Set Action Recognition}
\label{ssec:openset}

We now consider the open set action recognition problem, where we assume to have knowledge of $\cB$ known actions in the training stage while new or never-before-seen actions are also encountered in the test stage \cite{ScheirerRSB.PAMI2013,JainSB.ECCV2014,BendaleB.CVPR2015}. We first describe our approach for the open set problem based on the $k$ nearest neighbors classifier. Similar to the solution for the traditional multi-class closed set recognition problem described in Section~\ref{ssec:closedsetkNN}, we first compute distances between every pair of the training video sequences in each class using \eqref{eqn:sequencedist}. Then for every training sequence, we calculate the average distance between this training sequence and its $k$ nearest neighbors within the same class, and we use $\varphi_s$ to denote the maximum of all the averaged distances associated with all the training sequences in the $s$-th class ($s = 1, 2, \dots, \cB$). For a test sequence, we first apply $k$-NN to assign a ``tentative'' class membership for this sequence, which is denoted by $\widehat{s}$, and we use $\widehat{\varphi}$ to denote the average distance between this test sequence and its $k$ nearest training sequences in the $\widehat{s}$-th class. Finally, we declare this test sequence to be in the $\widehat{s}$-th class if $\widehat{\varphi} \leq \varphi_{\widehat{s}} \varsigma$, where $\varsigma > 1$ is a predefined thresholding parameter, otherwise it is labeled as a ``new'' action class.

Next, we discuss our solution for the open set problem based on one-vs.-all SVM. We first train $\cB$ one-vs.-all SVM classifiers using training sequences as proposed in Section~\ref{ssec:closedsetSVM}. Then for a test sequence, we declare it to be in the $\widehat{s}$-th class if ($i$) the classifier associated with the $\widehat{s}$-th action returns the maximum score among all the classifiers and ($ii$) this maximum score is above zero; otherwise, the test sequence is declared to be a ``new'' action.

\section{Experimental Results}
\label{sec:experiment}

We assess our proposed method on five human action datasets. Our first objective is to visually interpret and label the learned action attributes from HCS-LRR and to investigate the utility of the multi-resolution action attributes in semantic summarization of long video sequences of complex human activities. The second goal is to evaluate the performance of the learned action attributes in human action recognition/open set recognition and to compare our approach with several other union-of-subspaces learning methods.

\begin{figure}[t]
\centering
\subfigure[]{\includegraphics[height=0.6in]{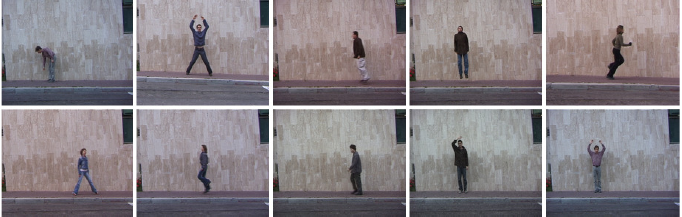} \label{fig:weizmanndata}}
\subfigure[]{\includegraphics[height=0.6in]{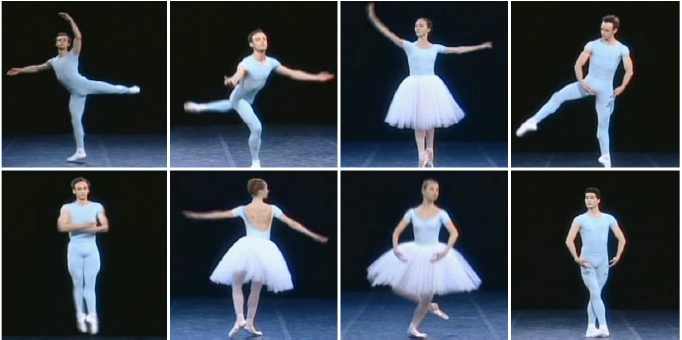} \label{fig:balletdata}}
\subfigure[]{\includegraphics[height=0.6in]{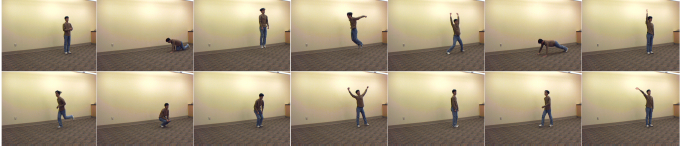} \label{fig:uiucdata}}
\subfigure[]{\includegraphics[height=0.6in]{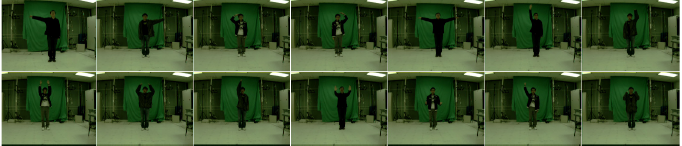} \label{fig:keckdata}}
\subfigure[]{\includegraphics[height=0.6in]{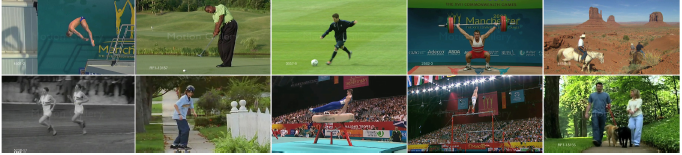} \label{fig:ucfdata}}
\caption{Sample frames from the five action recognition datasets used in our experiments: (a) Weizmann dataset; (b) Ballet dataset; (c) UIUC dataset; (d) Keck dataset and (e) UCF Sports dataset.}
\label{fig:dataplot}
\end{figure}

\subsection{Datasets}
\label{ssec:data}

The \emph{Weizmann} dataset \cite{GorelickBSIB.PAMI2007} consists of 90 low resolution ($180 \times 144$) video sequences from nine subjects. Each of the subjects performs 10 different actions: walk, run, jump, gallop sideways, bend, one-hand wave, two-hands wave, jump in place, jumping jack, and skip. The camera setting is fixed in this dataset. Sample frames are shown in Fig.~\ref{fig:weizmanndata}.

The \emph{Ballet} dataset \cite{FathiM.CVPR2008} contains 44 video sequences of 8 unique actions. The eight actions performed by three subjects are left-to-right hand opening, right-to-left hand opening, standing hand opening, leg swinging, jumping, turning, hopping, and standing still. Fig.~\ref{fig:balletdata} presents the sample frames of each action. Each video may contain several actions in this dataset. In total, there are 59 video clips, each containing only one action.

The \emph{UIUC} dataset \cite{TranS.ECCV2008} consists of 532 videos corresponding to eight subjects. In total, there are 14 action classes, including walking, running, jumping, waving, jumping jacks, clapping, jump from situp, raise one hand, stretching out, turning, sitting to standing, crawling, pushing up and standing to sitting (refer to Fig.~\ref{fig:uiucdata}). In this paper, we use the last section of the dataset, which contains 70 action sequences of all 14 actions performed by only one person and each action is performed five times.

The \emph{Keck} gesture dataset \cite{LinJD.ICCV2009} is collected using a camera with $640 \times 480$ resolution. It consists of 14 different gestures, including turn left, turn right, attention left, attention right, flap, stop left, stop right, stop both, attention both, start, go back, close distance, speed up, and come near. Example frames of this dataset are shown in Fig.~\ref{fig:keckdata}. Each of these 14 actions is performed by three people. In each video sequence, the subject repeats the same gesture three times. Therefore the total number of video sequences with static background is $14 \times 3 \times 3 = 126$.

The \emph{UCF Sports} dataset \cite{RodriguezAS.CVPR2008} contains 150 video samples, which are collected from various broadcast television channels such as BBC and ESPN. There are 10 action classes included in this dataset: diving, golf swing, kicking, lifting, horse riding, running, skateboarding, swinging-bench, swinging-side and walking, as shown in Fig.~\ref{fig:ucfdata}. In this paper, we exclude the videos belonging to horse riding and skateboarding actions as there is little to no body movement in these two classes. We also exclude two videos in the diving action since there are two subjects in each of those videos. This leaves us with 124 video sequences of 8 actions.

\subsection{Semantic Labeling and Summarization}
\label{ssec:semanticexperi}

\begin{figure*}[htbp]
\centering
\subfigure[\scriptsize Stand/Clapping]{\includegraphics[width=0.85in]{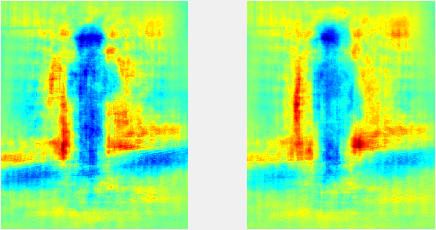}} \quad
\subfigure[Knee bend]{\includegraphics[width=0.85in]{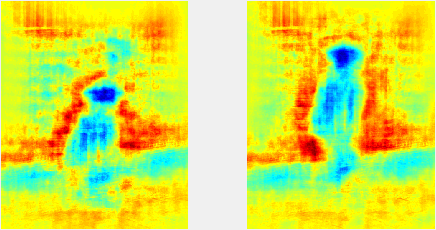}} \quad
\subfigure[\scriptsize Wave/Jumping jack]{\includegraphics[width=0.85in]{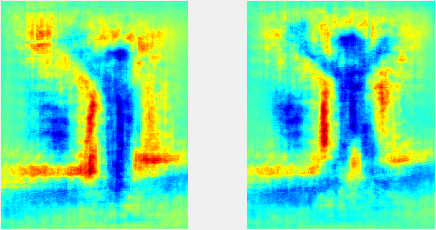}} \quad
\subfigure[\scriptsize Stretching out/Stand with one hand up]{\includegraphics[width=0.85in]{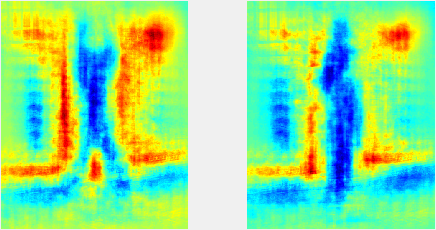}} \quad
\subfigure[\scriptsize Upper position of push up]{\includegraphics[width=0.85in]{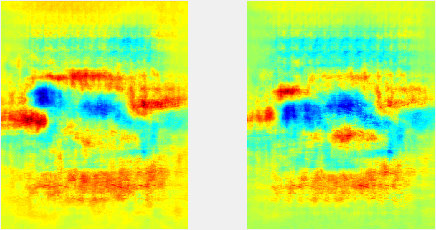}} \quad
\subfigure[\scriptsize Lower position of push up]{\includegraphics[width=0.85in]{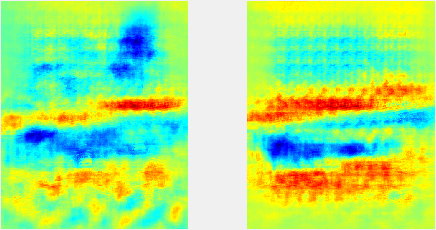}} \\
\subfigure[\scriptsize Stand/Clapping]{\includegraphics[width=0.85in]{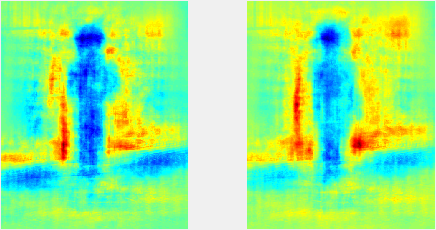}} \quad
\subfigure[Knee bend]{\includegraphics[width=0.85in]{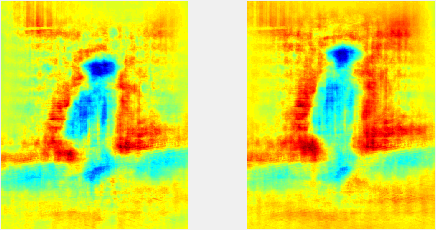}} \quad
\subfigure[\scriptsize Jumping jack/Stand with two hands out]{\includegraphics[width=0.85in]{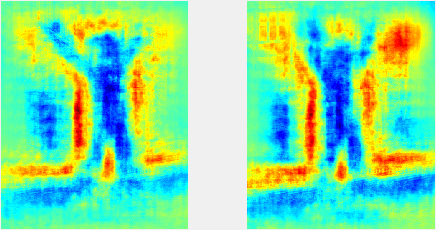}} \quad
\subfigure[Wave]{\includegraphics[width=0.85in]{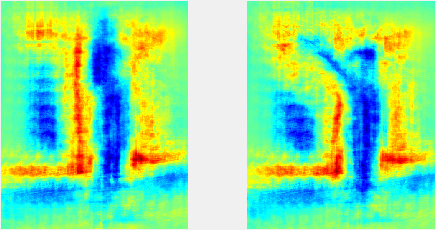}} \quad
\subfigure[\scriptsize Stand with one hand up]{\includegraphics[width=0.85in]{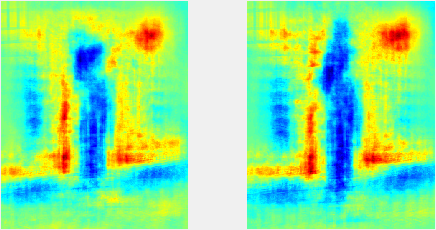}} \quad
\subfigure[\scriptsize Stretching out/Stand with two hands up]{\includegraphics[width=0.85in]{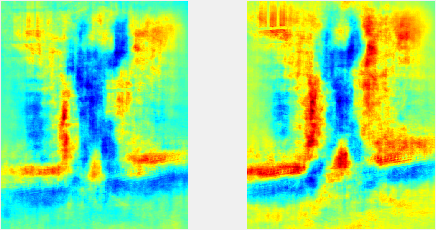}} \\
\subfigure[\scriptsize Upper position of push up]{\includegraphics[width=0.85in]{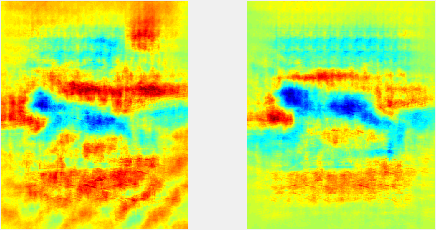}} \quad
\subfigure[\scriptsize Lower position of push up]{\includegraphics[width=0.85in]{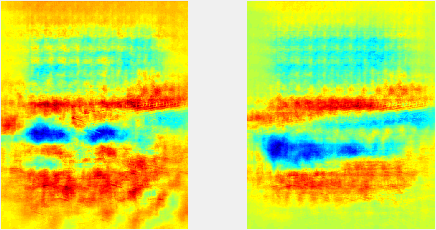}} \quad
\subfigure[\scriptsize Stand/Clapping]{\includegraphics[width=0.85in]{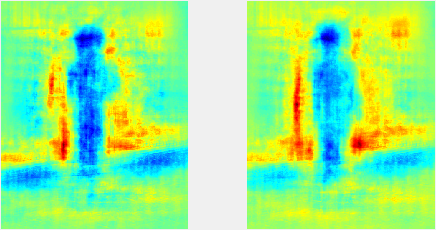}} \quad
\subfigure[\scriptsize Lower position of push up]{\includegraphics[width=0.85in]{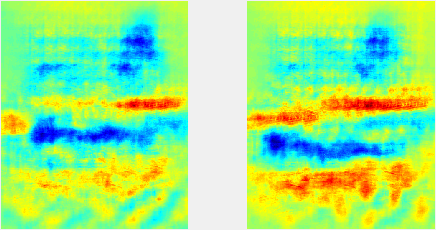}} \quad
\subfigure[Knee bend]{\includegraphics[width=0.85in]{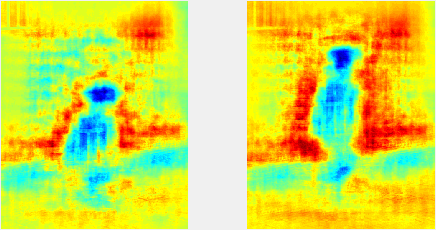}} \quad
\subfigure[Jumping jack]{\includegraphics[width=0.85in]{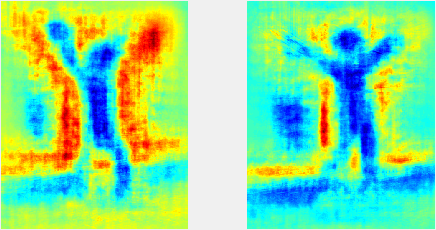}} \\
\subfigure[\scriptsize Upper position of push up]{\includegraphics[width=0.85in]{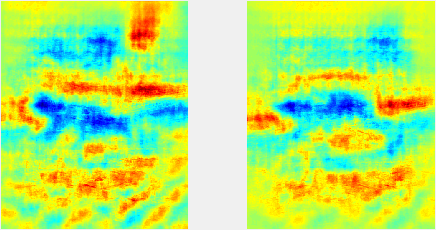}} \quad
\subfigure[Wave]{\includegraphics[width=0.85in]{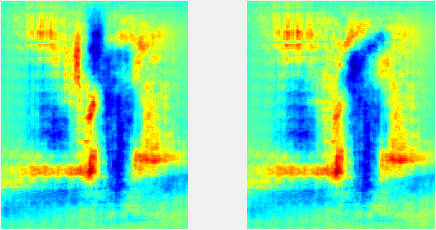}} \quad
\subfigure[\scriptsize Stand with one hand up/Stand]{\includegraphics[width=0.85in]{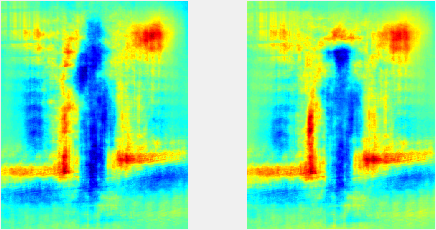}} \quad
\subfigure[\scriptsize Wave/Jumping jack]{\includegraphics[width=0.85in]{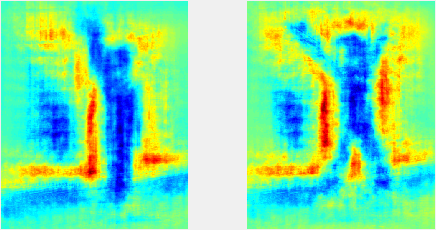}}
\caption{Visualization and interpretation of attributes learned using the frames from UIUC dataset. (a)-(f) represent the subspaces at the 3rd level of HCS-LRR. (g)-(n) represent the leaf subspaces (4th level) learned by HCS-LRR. (o)-(v) represent the subspaces which are learned using SC-LRR.}
\label{fig:uiucsubs}
\end{figure*}

\begin{figure*}[htbp]
\centering
\includegraphics[width=0.8\textwidth]{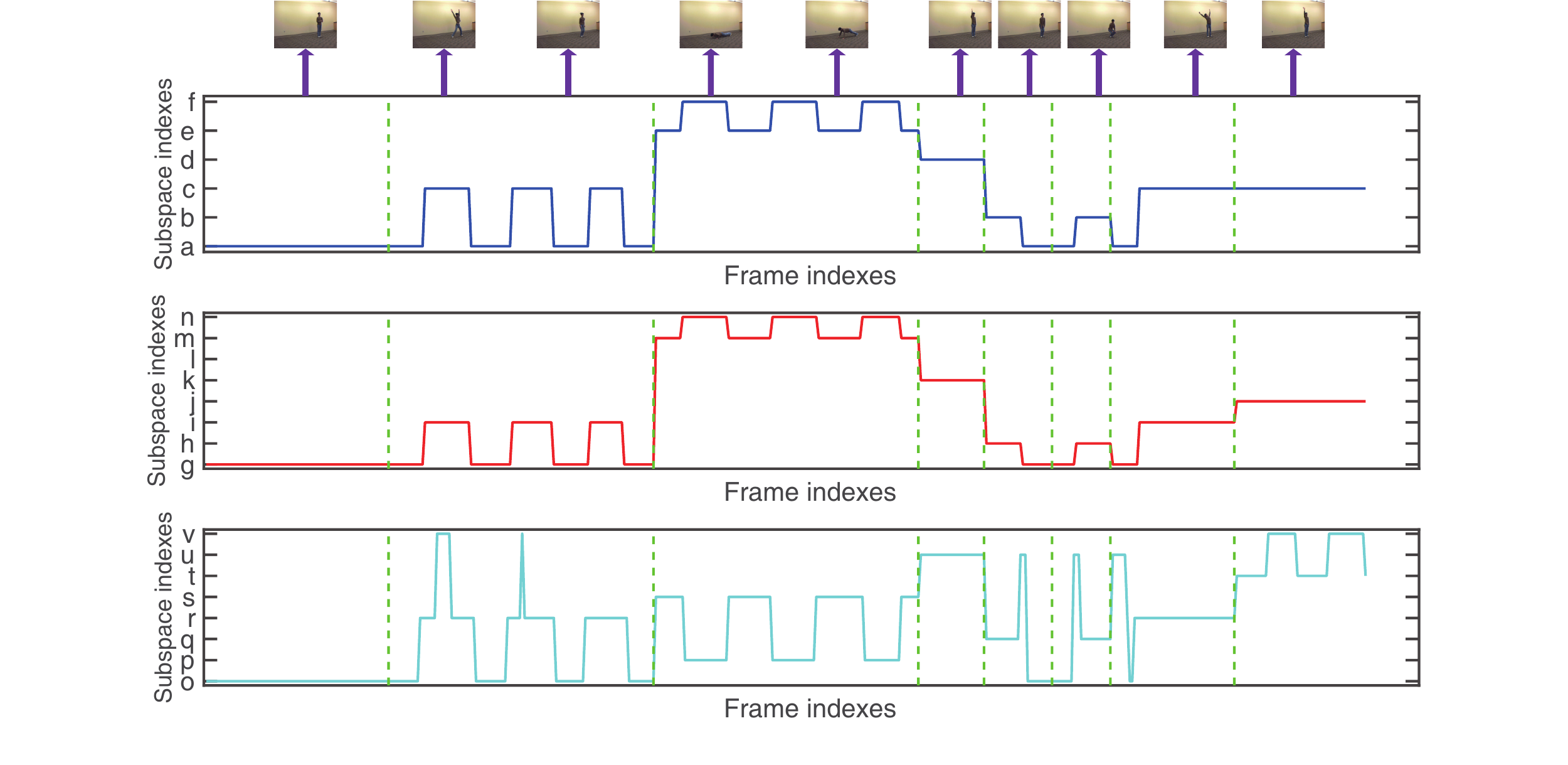}
\caption{Subspace transition of a long sequence using subspaces learned from the 3rd level of HCS-LRR (top), the bottom (4th) level of HCS-LRR (middle), and SC-LRR (bottom). The subspace assignment indexes correspond to the attributes in Fig.~\ref{fig:uiucsubs}.}
\label{fig:uiucsemantic}
\end{figure*}

In this section, we study the performance of HCS-LRR using HOG/MBH features for semantic summarization of multiple actions in a long video sequence. We compare the performance of HCS-LRR with several state-of-the-art subspace clustering algorithms such as Low-Rank Representation (LRR) \cite{LiuLYSYM.PAMI2013}, Sparse Subspace Clustering (SSC) \cite{ElhamifarV.PAMI2013}, Structure-Constrained LRR (SC-LRR) \cite{TangLSZ.TNNLS2014}, and Least Square Regression (LSR) \cite{LuMZZHY.ECCV2012}. The number of clusters $L$ for these methods is set to be the same number of subspaces generated by HCS-LRR at the bottom-most level. We first focus on UIUC dataset, where we select the first four video sequences of clapping, jumping jacks, pushing up, raise one hand, sitting to standing, standing to sitting, stretching out and waving actions and use the HOG features of these video sequences to learn the attributes. To extract HOG features, we use the bounding boxes provided by the authors of \cite{TranS.ECCV2008} and crop all the sequences into $544 \times 448$ aligned frames; then we set $n_{\sigma} = 32$ and hence $m = 2142$. We apply HCS-LRR with parameters $\alpha = 1.6$, $\beta = 0.5$, $\lambda = 0.5$, $P = 4$, $\gamma = 0.98$, $\varrho = 0.01$ and $d_{\mathrm{min}} = 3$. HCS-LRR returns 6 subspaces at the 3rd level and $L_P = 8$ leaf subspaces at the 4th level. In order to visualize these subspaces, we use patches of size $32 \times 32$ and contrast-sensitive HOG features (18-bin histogram) to learn the mapping between the HOG feature domain and the pixel domain, and coupled dictionaries with $K = 1000$ atoms are learned using \eqref{eqn:dictlearn}. The first two dimensions of the bases of those subspaces in the 3rd and 4th level with their semantic labels are shown in Fig.~\ref{fig:uiucsubs} (a)-(f) and Fig.~\ref{fig:uiucsubs} (g)-(n), respectively. It can be seen that the attributes (a), (b), (e) and (f) from the 3rd level are leaf nodes and hence, translate as they were to the 4th level. However, the attribute (c) corresponding to Wave/Jumping jack at the 3rd level is further divided into attributes (i) and (j) corresponding to Jumping jack/Stand with two hands out and One-hand wave at the 4th level, respectively. Similarly, the attribute (d) corresponding to Stretching out/Stand with one hand up at the 3rd level is further divided into attributes (k) and (l) corresponding to Stand with one hand up and Stretching out/Stand with two hands up in the bottom level, respectively. To demonstrate semantic summarization using these learned attributes on a complex human activity, we select the last video sequence of each of the aforementioned eight actions as the test sequences and create a long video stream by concatenating these sequences. The semantic summary is illustrated as the subspace transition of the frames in Fig.~\ref{fig:uiucsemantic}, where different actions in the sequence are delimited by green lines. One can summarize the actions in terms of natural language by using the attribute assignment indexes and the corresponding labels. We compare HCS-LRR with other subspace clustering algorithms in terms of action recognition of the test sequences by setting $L = 8$ for other approaches. SC-LRR also gives perfect recognition accuracy based on $k$-NN classifier. To compare SC-LRR with our technique in terms of semantic summarization, we apply SC-LRR on the training sequences to learn 8 attributes, and the corresponding labeled subspaces are illustrated in Fig.~\ref{fig:uiucsubs} (o)-(v). However, compared to the attributes at the bottom level of HCS-LRR, we observe that each of the attributes (u) and (v) learned using SC-LRR can have two totally different interpretations. For instance, attribute (v) is a mixture of ``one-hand wave'' and ``jumping jack'' action attributes. There are two advantages of using HCS-LRR model over other techniques. First, the semantic summary of the video sequence can be generated at different resolutions using attributes at different levels of the hierarchical structure. Second, the summary generated using the attributes learned from HCS-LRR is very clear because the actions are well-clustered into different attributes, but it is difficult to generate a summary using the attributes learned from SC-LRR alone due to confusion in the last two attributes. We provide the evidence of these advantages in Fig.~\ref{fig:uiucsemantic}. In short, the semantic interpretation performance using HCS-LRR seems better compared to SC-LRR.

\begin{figure*}[htbp]
\centering
\subfigure[Arabesque]{\includegraphics[width=0.85in]{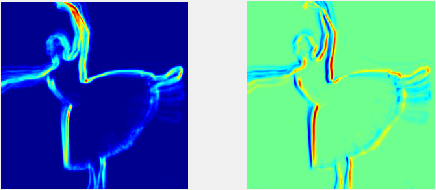}} \quad
\subfigure[Arabesque]{\includegraphics[width=0.85in]{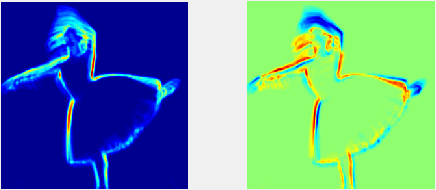}} \quad
\subfigure[D{\'e}velopp{\'e}]{\includegraphics[width=0.85in]{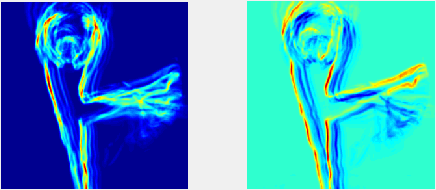}} \quad
\subfigure[Arabesque]{\includegraphics[width=0.85in]{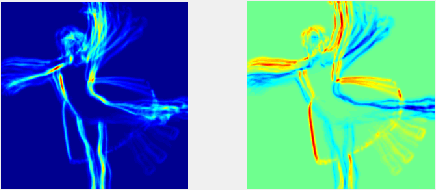}} \quad
\subfigure[Pirouette with working leg straight]{\includegraphics[width=0.85in]{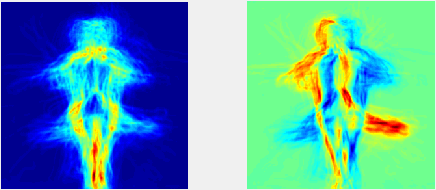}} \quad
\subfigure[{\`A} la seconde]{\includegraphics[width=0.85in]{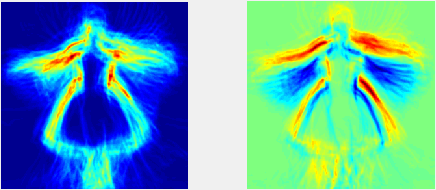}} \\
\subfigure[Changement]{\includegraphics[width=0.85in]{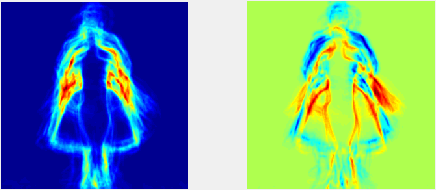}} \quad
\subfigure[Changement]{\includegraphics[width=0.85in]{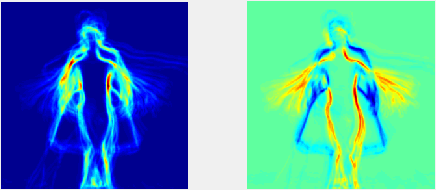}} \quad
\subfigure[Crois{\'e} devant]{\includegraphics[width=0.85in]{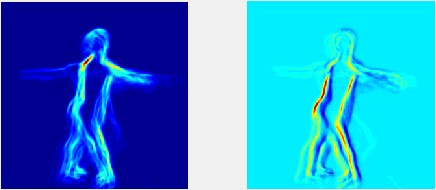}} \quad
\subfigure[Arabesque]{\includegraphics[width=0.85in]{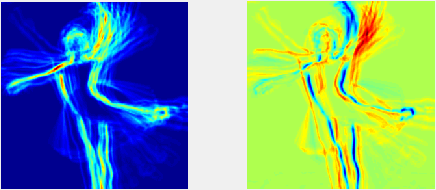}} \quad
\subfigure[{\`A} la seconde]{\includegraphics[width=0.85in]{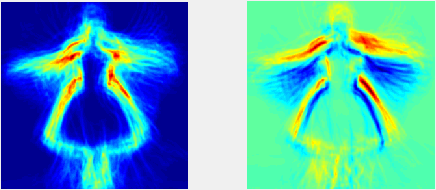}} \quad
\subfigure[D{\'e}velopp{\'e}]{\includegraphics[width=0.85in]{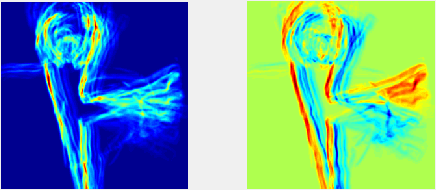}} \\
\subfigure[Pirouette with working leg straight]{\includegraphics[width=0.85in]{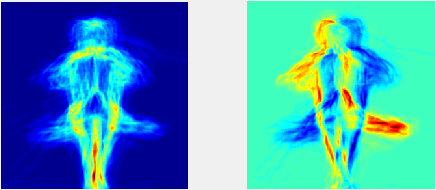}} \quad
\subfigure[Arabesque]{\includegraphics[width=0.85in]{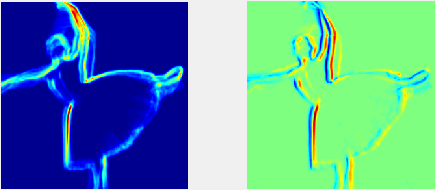}}
\caption{Visualization and interpretation of attributes learned using the frames from Ballet dataset. (a)-(g) represent the leaf subspaces learned by HCS-LRR. (h)-(n) represent the subspaces which are learned using SSC.}
\label{fig:balletsubs}
\end{figure*}

\begin{figure*}[htbp]
\centering
\includegraphics[width=0.8\textwidth]{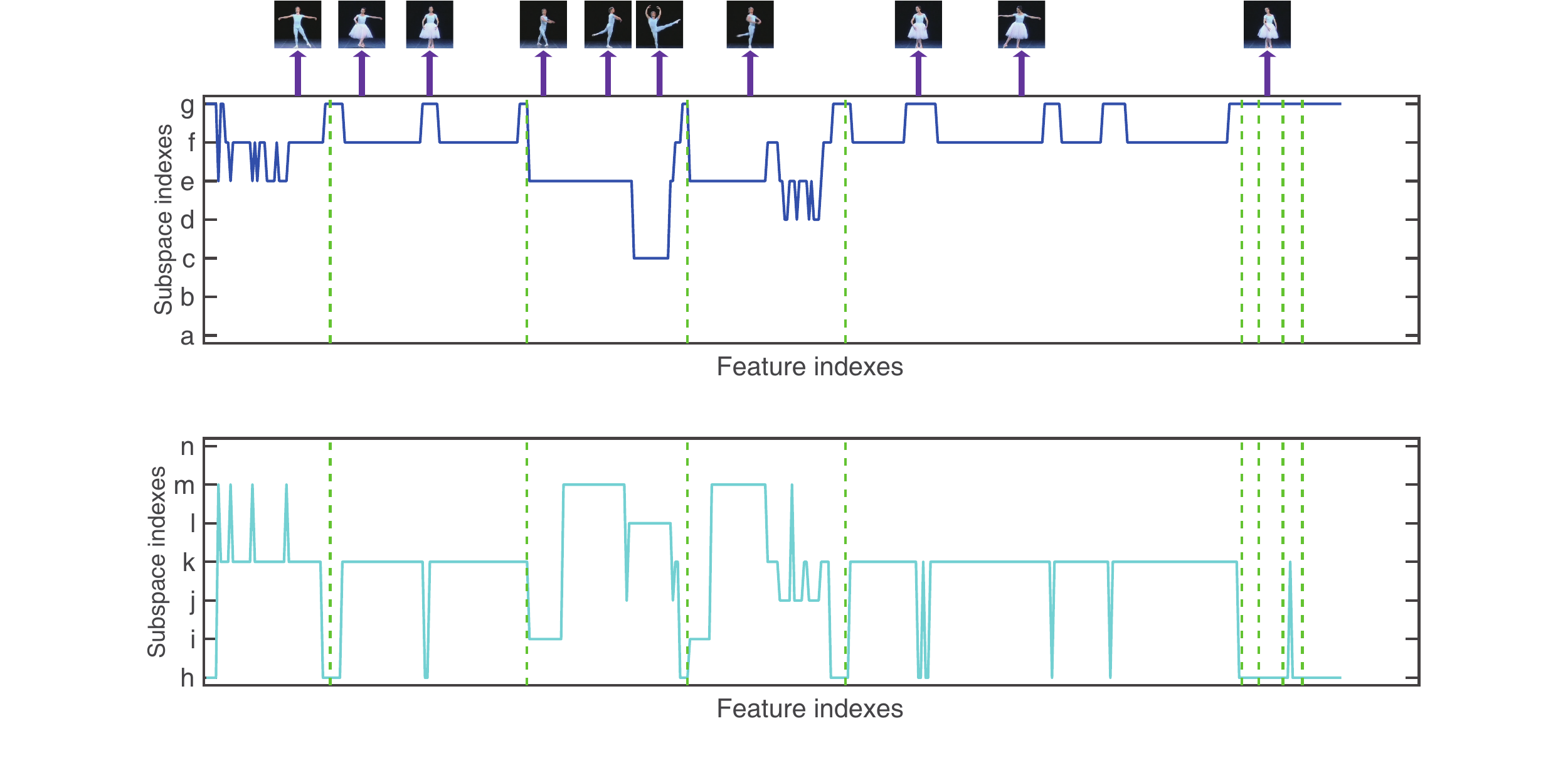}
\caption{Subspace transition of a long sequence using subspaces learned from the 3rd level of HCS-LRR (top) and SSC (bottom). The subspace assignment indexes correspond to the attributes in Fig.~\ref{fig:balletsubs}.}
\label{fig:balletsemantic}
\end{figure*}

As another example, we experiment with Ballet dataset, where we select video clips from standing hand opening, turning, hopping and standing still actions for learning the attributes, with the number of sequences set to be 5, 5, 2 and 6, respectively. We crop all the video frames into $288 \times 288$ pixels and extract MBH descriptors with $n_{\sigma} = 32$ and $n_{\tau} = 3$; thus $m = 1458$. Then we implement HCS-LRR with parameters $\alpha = 0.8$, $\beta = 0.5$, $\lambda = 0.4$, $P = 3$, $\gamma = 0.98$, $\varrho = 0.05$ and $d_{\mathrm{min}} = 3$. It has $L_P = 7$ leaf subspaces at the 3rd level, and the first two basis vectors of each leaf subspace are illustrated in Fig.~\ref{fig:balletsubs} (a)-(g). Optical flow frames are used for visualization as explained in Section~\ref{ssec:mbhvis}. For testing purpose, we again choose the video clips belonging to the aforementioned actions, with the number of clips set to be 2, 2, 1 and 4, respectively. For these 9 test sequences from these four different actions, attributes learned using SSC with $L = 7$ give perfect ($100\%$) action recognition performance when using $k$-NN classifier. The subspaces learned using SSC are shown in Fig.~\ref{fig:balletsubs} (h)-(n). We again concatenate all the test video clips to create a long video sequence and visualize the subspace transition of the MBH features of the long sequence in Fig.~\ref{fig:balletsemantic}. In the case of HCS-LRR, it can be observed that the attribute (b) does not provide any additional information than that is being provided by attribute (a). The attribute (i) learned using SSC is not seen among the attributes at the bottom level of HCS-LRR. In addition, there are no features of the test sequence that get assigned to attributes (a), (b) and (n) due to the significant intraclass variations. In other words, SSC seems to offer better summarization performance than HCS-LRR. However, the caveat here is that in practice, the number of subspaces or attributes is not known a priori for SSC, while HCS-LRR algorithm can automatically give an estimate of $L$.

\subsection{Action Recognition}
\label{ssec:classifyexperi}

To quantify the discriminative power of the subspaces learned from HCS-LRR, we also perform human action recognition using the action attributes learned by our proposed method. We again compare the performance of HCS-LRR with LRR \cite{LiuLYSYM.PAMI2013}, SSC \cite{ElhamifarV.PAMI2013}, SC-LRR \cite{TangLSZ.TNNLS2014}, and LSR \cite{LuMZZHY.ECCV2012}. Different from the experiments in Section~\ref{ssec:semanticexperi}, the number of clusters $L$ for these algorithms is now set ($i$) to be the same number of leaf subspaces generated by HCS-LRR (which is denoted by $\langle$\emph{Algorithm}$\rangle$-$L_P$) and ($ii$) to be the same as the number of actions in the training data (which is denoted by $\langle$\emph{Algorithm}$\rangle$-$\cB$). For all these methods, including HCS-LRR, we first tune the parameters for clustering to achieve their best recognition performance based on $k$-NN classifier. Once the attributes are learned using these parameters, we further tune the parameters for SVM classifiers to obtain their best recognition performance. We simply fix $\gamma = 0.98$ for all the experiments. It should be noted that ``SVM/ova'' and ``SVM/ovo'' in Table~\ref{tab:closedsetclassify} stand for ``one-vs.-all SVM'' and ``one-vs.-one SVM,'' respectively.

\subsubsection{Closed Set Recognition}
\label{sssec:closedsetexperi}

\begin{table*}[htbp]
\centering
\caption{Closed Set Action Recognition Results ($\%$)}
\begin{tabular}{c||c|c|c|c|c|c|c|c|c|c|c}
\hline
\multirow{2}{*}{Dataset} & \multirow{2}{*}{Feature} & \multirow{2}{*}{Classifier} & \multicolumn{9}{c}{Subspace clustering method}  \\
\cline{4-12}
  & & & HCS-LRR & LRR-$L_P$ & LRR-$\cB$ & SSC-$L_P$ & SSC-$\cB$ & SC-LRR-$L_P$ & SC-LRR-$\cB$ & LSR-$L_P$ & LSR-$\cB$  \\
\hline
\multirow{6}{*}{Weizmann} & \multirow{3}{*}{HOG} & $k$-NN & \textbf{90.00} & 73.33 & 67.78 & 58.89 & 52.22 & 76.67 & 66.67 & 67.78 & 67.78  \\
\cline{3-12}
 & & SVM/ova & \textbf{92.22} & 65.56 & 52.22 & 58.89 & 48.89 & 64.44 & 63.33 & 58.89 & 65.56  \\
\cline{3-12}
 & & SVM/ovo & \textbf{95.56} & 82.22 & 66.67 & 64.44 & 51.11 & 68.89 & 68.89 & 71.11 & 72.22  \\
\cline{2-12}
 & \multirow{3}{*}{MBH} & $k$-NN & \textbf{91.11} & 88.89 & 66.67 & 86.67 & 65.56 & 90.00 & 72.22 & 90.00 & 73.33  \\
\cline{3-12}
 & & SVM/ova & \textbf{87.78} & 85.56 & 64.44 & 85.56 & 64.44 & 81.11 & 68.89 & 86.67 & 71.11  \\
\cline{3-12}
 & & SVM/ovo & 85.56 & \textbf{87.78} & 66.67 & 84.44 & 61.11 & 81.11 & 66.67 & 85.56 & 71.11  \\
\hline
\hline
\multirow{6}{*}{Ballet} & \multirow{3}{*}{HOG} & $k$-NN & \textbf{71.19} & 54.24 & 32.20 & 52.54 & 38.98 & 45.76 & 49.15 & 40.68 & 45.76  \\
\cline{3-12}
 & & SVM/ova & \textbf{67.80} & 59.32 & 30.51 & \textbf{67.80} & 38.98 & 66.10 & 42.37 & 54.24 & 61.02  \\
\cline{3-12}
 & & SVM/ovo & 62.71 & 62.71 & 52.54 & \textbf{64.41} & 42.37 & \textbf{64.41} & 57.63 & 54.24 & 61.02  \\
\cline{2-12}
 & \multirow{3}{*}{MBH} & $k$-NN & \textbf{69.49} & 61.02 & 54.24 & 57.63 & 50.85 & 57.63 & 54.24 & 61.02 & 52.54  \\
\cline{3-12}
 & & SVM/ova & \textbf{71.19} & 69.49 & 61.02 & 54.24 & 20.34 & 61.02 & 62.71 & 67.80 & 67.80  \\
\cline{3-12}
 & & SVM/ovo & \textbf{69.49} & 67.80 & 61.02 & 45.76 & 44.07 & 67.80 & 61.02 & 62.71 & 67.80  \\
\hline
\hline
\multirow{6}{*}{UIUC} & \multirow{3}{*}{HOG} & $k$-NN & \textbf{100} & 94.29 & 98.57 & 98.57 & 85.71 & \textbf{100} & 95.71 & 98.57 & 95.71  \\
\cline{3-12}
 & & SVM/ova & \textbf{100} & 77.14 & 91.43 & 98.57 & 81.43 & \textbf{100} & 90.00 & \textbf{100} & 91.43  \\
\cline{3-12}
 & & SVM/ovo & \textbf{100} & 92.86 & 98.57 & \textbf{100} & 78.57 & \textbf{100} & 91.43 & \textbf{100} & 85.71  \\
\cline{2-12}
 & \multirow{3}{*}{MBH} & $k$-NN & \textbf{100} & \textbf{100} & 98.57 & \textbf{100} & 98.57 & \textbf{100} & 82.86 & \textbf{100} & 98.57  \\
\cline{3-12}
 & & SVM/ova & \textbf{100} & \textbf{100} & 95.71 & \textbf{100} & \textbf{100} & \textbf{100} & 78.57 & \textbf{100} & \textbf{100}  \\
\cline{3-12}
 & & SVM/ovo & \textbf{100} & \textbf{100} & 95.71 & \textbf{100} & \textbf{100} & \textbf{100} & 77.14 & \textbf{100} & \textbf{100}  \\
\hline
\hline
\multirow{3}{*}{Keck} & \multirow{3}{*}{MBH} & $k$-NN & \textbf{88.10} & 83.33 & 78.57 & 79.37 & 83.33 & 80.16 & 80.95 & 76.19 & 70.63  \\
\cline{3-12}
 & & SVM/ova & \textbf{87.30} & 76.98 & 66.67 & 74.60 & 71.43 & 79.37 & 80.16 & 57.14 & 61.11  \\
\cline{3-12}
 & & SVM/ovo & \textbf{90.48} & 76.98 & 74.60 & 78.57 & 73.81 & 84.92 & 84.13 & 70.63 & 69.05  \\
\hline
\hline
\multirow{3}{*}{UCF} & \multirow{3}{*}{MBH} & $k$-NN & \textbf{57.26} & 50.81 & 47.58 & 56.45 & 47.58 & 45.97 & 38.71 & 55.65 & 43.55  \\
\cline{3-12}
 & & SVM/ova & 66.13 & \textbf{67.74} & 60.48 & 66.13 & 54.84 & 64.52 & 50.81 & 64.52 & 54.03  \\
\cline{3-12}
 & & SVM/ovo & \textbf{75.81} & 75.00 & 57.26 & \textbf{75.81} & 53.23 & 73.39 & 45.16 & 68.55 & 50.00  \\
\hline
\end{tabular}
\label{tab:closedsetclassify}
\end{table*}

In this subsection, we carry out the closed set action recognition experiments on the five datasets described in Section~\ref{ssec:data}. We begin our experiments on Weizmann dataset, where we use both HOG and MBH features to learn the attributes and evaluate all the subspace/attribute learning approaches based on a leave-one-subject-out experiment. We crop all the sequences into $88 \times 64$ aligned frames and HOG features are extracted with blocks of size $n_{\sigma} = 8$, resulting in $m = 792$. Then we perform HCS-LRR with parameters $\alpha = 1.2$, $\beta = 0.8$, $\lambda = 0.8$, $P = 5$, $\varrho = 0.01$ and $d_{\mathrm{min}} = 5$, in which setting it returns $L_P = 24$ leaf subspaces for final attributes. As shown in Table~\ref{tab:closedsetclassify}, our method significantly outperforms other subspace clustering approaches for all the classifiers. To be specific, the lowest error is achieved by HCS-LRR when used with one-vs.-one SVM classifier and the resulting recognition accuracy is $95.56\%$. Similarly, we crop all the optical flow fields into $88 \times 64$ images and extract MBH features with $n_{\sigma} = 8$ and $n_{\tau} = 3$; hence $m = 1584$. We apply HCS-LRR with parameters $\alpha = 0.4$, $\beta = 0.7$, $\lambda = 0.4$, $P = 5$, $\varrho = 0.01$ and $d_{\mathrm{min}} = 12$, obtaining $L_P = 23$ final subspaces. We can see from Table~\ref{tab:closedsetclassify} that for a fixed classifier, the recognition accuracy of all the approaches is close to each other for 23 clusters. Our algorithm achieves the highest classification accuracy among all the methods for $k$-NN classifier and one-vs.-all SVM classifier. When using one-vs.-one SVM classifier, LRR-$L_P$ outperforms others and HCS-LRR is only $2.2\%$ (corresponds to 2 samples) behind LRR-$L_P$. The main reason for $k$-NN classifier outperforming SVM classifiers is that the original feature vectors are used for aligning the training and test sequences in the case of $k$-NN classification, and one can get perfect alignment of the video sequences using MBH features on clean datasets such as Weizmann dataset. However, the SVM classifiers only use subspace transition vectors instead of using the original feature vectors for alignment of the sequences via the DTW kernel, which may degrade the recognition performance.

For Ballet dataset, we crop all the video frames into $288 \times 288$ pixels and extract HOG descriptor of every frame with $n_{\sigma} = 32$; hence $m = 729$. Since there is no significant motion between consecutive frames, instead of using HOG features of each frame separately, we take the sum of the HOG features of two adjacent frames at a time and the sum is used as the feature of two adjacent frames. We perform HCS-LRR with parameters $\alpha = 1.5$, $\beta = 0.2$, $\lambda = 2.2$, $P = 4$, $\varrho = 0.05$ and $d_{\mathrm{min}} = 3$. The number of subspaces returned by HCS-LRR at the finest scale is $L_P = 11$. We evaluate the subspace learning approaches using the leave-one-sequence-out scheme. The results presented in Table~\ref{tab:closedsetclassify} show that the proposed algorithm generates the lowest error rates for $k$-NN and one-vs.-all SVM classifiers. Both SSC-$L_P$ and SC-LRR-$L_P$ perform the best for one-vs.-one SVM classifier, and our method is only 1 sample lower than these two approaches. After extracting the MBH descriptors with the same parameters as in Section~\ref{ssec:semanticexperi}, we set $\alpha = 0.6$, $\beta = 0.2$, $\lambda = 0.2$, $P = 4$, $\varrho = 0.05$ and $d_{\mathrm{min}} = 3$ for HCS-LRR, which gives us $L_P = 9$ leaf subspaces. The results are listed in Table~\ref{tab:closedsetclassify}, from which we make the conclusion that by representing the human actions using the attributes learned by HCS-LRR, we are able to recognize the actions at a superior rate compared to other techniques.

Next, we evaluate our approach on UIUC dataset, where we again use both HOG and MBH features and conduct leave-one-sequence-out experiments. To extract HOG features, we crop all the sequences into $544 \times 448$ aligned frames by using the bounding boxes provided by the authors of \cite{TranS.ECCV2008}, and follow the same parameter setting as in Section~\ref{ssec:semanticexperi}. We perform HCS-LRR with parameters $\alpha = 1$, $\beta = 0.2$, $\lambda = 0.1$, $P = 5$, $\varrho = 0.01$ and $d_{\mathrm{min}} = 3$, in which case it returns $L_P = 18$ leaf subspaces. We crop all the optical flow fields and the action interest region also has $544 \times 448$ pixels, then MBH features are extracted with $n_{\sigma} = 32$ and $n_{\tau} = 2$; therefore $m = 4284$ in this setting. We set $\alpha = 0.2$, $\beta = 0.1$, $\lambda = 0.1$, $P = 5$, $\varrho = 0.01$ and $d_{\mathrm{min}} = 5$ for HCS-LRR and finally there are $L_P = 26$ subspaces at the bottom-most level. The recognition accuracy for all the methods is reported in Table~\ref{tab:closedsetclassify}. As Table~\ref{tab:closedsetclassify} shows, both our method and SC-LRR-$L_P$ obtain $100\%$ recognition accuracy for both HOG and MBH features. In general, using the number of clusters that is automatically generated from HCS-LRR will improve the performance for almost all the methods.

To evaluate different methods on Keck dataset, we employ the leave-one-person-out cross validation protocol. We only use MBH features to learn the attributes on this dataset since the background of the aligned sequences is dynamic. We crop all the optical flow fields into $288 \times 288$ images and extract MBH features with $n_{\sigma} = 32$ and $n_{\tau} = 3$, thereby resulting in $m = 1458$. We implement HCS-LRR with parameters $\alpha = 0.1$, $\beta = 0.1$, $\lambda = 0.2$, $P = 5$, $\varrho = 0.01$ and $d_{\mathrm{min}} = 5$. The number of leaf attributes returned by HCS-LRR is $L_P = 28$. The classification results are summarized in Table~\ref{tab:closedsetclassify}. Our approach shows noticeable improvement in accuracy compared to other state-of-the-art subspace clustering algorithms. The recognition rate reaches $90.48\%$ for HCS-LRR coupled with one-vs.-one SVM classifier, which outperforms SC-LRR-$L_P$, the second best, by $5.5\%$.

Finally, we examine the performance of different methods on UCF Sports dataset. We align the sequences by applying the method proposed in \cite{ZhouYY.PAMI2013} and subtract the background using the approach described in \cite{YeYSLHW.CSVT2015}, then we use the bounding boxes provided by the authors of \cite{RodriguezAS.CVPR2008} to crop all the sequences and reshape them into $416 \times 256$ frames. Due to the dynamic property of the background, we again only use MBH features to learn the attributes, where the MBH features are extracted with $n_{\sigma} = 32$ and $n_{\tau} = 3$; so that $m = 1872$. We investigate the performance of the subspace clustering approaches via the leave-one-sequence-out scheme. We use the following parameters for HCS-LRR: $\alpha = 1.2$, $\beta = 0.1$, $\lambda = 0.3$, $P = 5$, $\varrho = 0.005$ and $d_{\mathrm{min}} = 3$. Finally, the output $L_P$ from HCS-LRR is 26 in this setting. Table~\ref{tab:closedsetclassify} presents the recognition results for all the methods. For $k$-NN classifier, our algorithm achieves the highest classification accuracy, with 1 sample higher than the second best, SSC-$L_P$. The classification performance improves drastically when using SVM classifiers. For one-vs.-one SVM classifier, both HCS-LRR and SSC-$L_P$ achieve $75.81\%$ classification accuracy, but HCS-LRR method has an additional advantage that it can generate the number of clusters automatically without any prior knowledge of the data.

\subsubsection{Open Set Recognition}
\label{sssec:opensetexperi}

\begin{table*}[htbp]
\centering
\caption{Open Set Action Recognition Results ($\%$)}
\begin{tabular}{c||c|c|c|c|c|c|c|c|c|c}
\hline
\multirow{2}{*}{Dataset} & \multirow{2}{*}{Classifier} & \multicolumn{9}{c}{Subspace clustering method}  \\
\cline{3-11}
  & & HCS-LRR & LRR-$L_P$ & LRR-$\cB$ & SSC-$L_P$ & SSC-$\cB$ & SC-LRR-$L_P$ & SC-LRR-$\cB$ & LSR-$L_P$ & LSR-$\cB$  \\
\hline
\multirow{2}{*}{UIUC} & $k$-NN & \textbf{100} & 90.91 & 86.36 & 90.91 & 86.36 & 95.45 & 95.45 & 90.91 & 86.36  \\
\cline{2-11}
 & one-vs.-all SVM & \textbf{100} & 95.45 & 90.91 & 90.91 & 90.91 & \textbf{100} & 86.36 & 95.45 & 90.91  \\
\hline
\hline
\multirow{2}{*}{Keck} & $k$-NN & \textbf{89.58} & 79.17 & 77.08 & 79.17 & 75.00 & 72.92 & 81.25 & 77.08 & 75.00  \\
\cline{2-11}
 & one-vs.-all SVM & \textbf{89.58} & 83.33 & 85.42 & 79.17 & 77.08 & 85.42 & 87.50 & 81.25 & 81.25  \\
\hline
\end{tabular}
\label{tab:opensetclassify}
\end{table*}

One important advantage of learning human action attributes instead of representing an entire action as a single feature vector is that an action that is absent in the training stage can be identified as a new action class if it can be represented by the attributes learned from other actions in the training data. Here, we show the effectiveness of HCS-LRR for the open set action recognition problem. Consider a collection of $N_k$ test samples from $\cB$ known classes and $N_u$ test samples from $\cU$ unknown classes, which are denoted by $\{ (\mathbf{\Phi}_i, \vartheta_i) \}_{i=1}^{N_k+N_u}$, where $\vartheta_i \in \{ 1, 2, \dots, \cB \}$ for $i \leq N_k$ and $\vartheta_i = new$ for $i > N_k$. We can measure the performance of our algorithm under the classification rule $\cC$ as $\epsilon_o = \frac{\sum_{i=1}^{N_k+N_u} [\cC(\mathbf{\Phi}_i) = \vartheta_i]}{N_k+N_u}$.

UIUC dataset is used for the first set of experiments. From the 70 sequences in the dataset, we select the first four video sequences of all classes except ``walking'' and ``standing to sitting'' actions for training and the remaining 22 sequences for testing; thus $\cB = 12$, $\cU = 2$, $N_k = 12$ and $N_u = 10$. We extract HOG features of the training sequences as in Section~\ref{sssec:closedsetexperi} and perform HCS-LRR with parameters $\alpha = 0.8$, $\beta = 0.1$, $\lambda = 0.4$, $P = 5$, $\varrho = 0.01$ and $d_{\mathrm{min}} = 3$, which in turn provides us with $L_P = 19$ leaf subspaces. We can observe from Table~\ref{tab:opensetclassify} that HCS-LRR achieves perfect accuracy for both classifiers. Moreover, the recognition accuracy is improved when we set $L = 19$ instead of 12 for other subspace clustering approaches.

The second set of experiments are performed on Keck dataset. We select the video sequences associated with the last two subjects from all classes except ``come near'' action for training and the remaining 48 sequences for testing; hence $\cB = 13$, $\cU = 1$, $N_k = 39$ and $N_u = 9$. We extract MBH features of the training sequences as in Section~\ref{sssec:closedsetexperi} and set $\alpha = 0.1$, $\beta = 0.9$, $\lambda = 0.8$, $P = 5$, $\varrho = 0.01$ and $d_{\mathrm{min}} = 5$ for HCS-LRR. There are $L_P = 18$ attributes learned from HCS-LRR at the bottom level. Table~\ref{tab:opensetclassify} summarizes the $\epsilon_o$'s of all the algorithms for both classifiers. As can be seen, HCS-LRR outperforms other subspace clustering algorithms, which again proves the effectiveness of our method. In addition, the performance of other subspace clustering algorithms can be significantly enhanced when we use one-vs.-all SVM for recognition.

\section{Conclusion}
\label{sec:conclude}

In this paper, we proposed a novel extension of the canonical low-rank representation (LRR), termed the clustering-aware structure-constrained low-rank representation (CS-LRR) model, for unsupervised data-driven human action attribute learning. The proposed CS-LRR model incorporates spectral clustering into the learning framework, which helps spectral clustering achieve the best clustering results. Moreover, we introduced a hierarchical subspace clustering approach based on CS-LRR model, called HCS-LRR model, which does not need the number of clusters to be specified a priori. Experimental results on real video datasets showed the effectiveness of our clustering approach and its superiority over the state-of-the-art union-of-subspaces learning algorithms for its utility in semantic summarization of complex human activities at multiple resolutions as well as human action recognition.

\end{document}